\documentclass{article}

\PassOptionsToPackage{numbers, compress}{natbib}
\usepackage[final]{neurips_2024}

\usepackage[utf8]{inputenc} % allow utf-8 input
\usepackage[T1]{fontenc}    % use 8-bit T1 fonts
\usepackage{hyperref}       % hyperlinks
\usepackage{url}            % simple URL typesetting
\usepackage{booktabs}       % professional-quality tables
\usepackage{amsfonts}       % blackboard math symbols
\usepackage{nicefrac}       % compact symbols for 1/2, etc.
\usepackage{microtype}      % microtypography
\usepackage{lipsum}  
\usepackage{kotex}
\usepackage{algorithm}
\usepackage{algpseudocode}
\usepackage{amsmath}
\usepackage{amssymb}
\usepackage{amsthm}
\usepackage[utf8]{inputenc}
\usepackage{graphics}
\usepackage{amsmath,amsfonts,bm}

\newcommand{\RNum}[1]{\uppercase\expandafter{\romannumeral #1\relax}}
\newcommand{\Rnum}[1]{\lowercase\expandafter{\romannumeral #1\relax}}

\def\Secref#1{Section~\ref{#1}}
\def\Figref#1{Fig.~\ref{#1}}
\def\Tabref#1{Table~\ref{#1}}

\def\Secref#1{Section~\ref{#1}}
\def\eqref#1{(\textcolor{red}{\ref{#1}})}
\def\0{\bm{0}} %%% LSY
\def\1{\bm{1}}

\def\rvp{{\mathbf{p}}}
\def\rvq{{\mathbf{q}}}

\def\rvx{{\mathbf{x}}}

\def\vy{{\bm{y}}}

\DeclareMathAlphabet{\mathsfit}{\encodingdefault}{\sfdefault}{m}{sl}
\SetMathAlphabet{\mathsfit}{bold}{\encodingdefault}{\sfdefault}{bx}{n}

\usepackage{graphicx}
\usepackage{subfig}
\usepackage{wrapfig}
\usepackage{bbm}
\usepackage[table,xcdraw]{xcolor}

\usepackage{makecell}
\usepackage{multirow}
\usepackage{soul}
\usepackage{mathtools}
\usepackage{rotating}
\usepackage{afterpage}

\theoremstyle{definition}

\renewcommand{\thefootnote}{\fnsymbol{footnote}}
\title{Are Self-Attentions Effective \\for Time Series Forecasting?}

\author{%
  Dongbin Kim\textsuperscript{1},\quad Jinseong Park\textsuperscript{1},\quad Jaewook Lee\textsuperscript{1*},\quad Hoki Kim\textsuperscript{2*} \\[1ex]
  \textsuperscript{1}Seoul National University  \quad \quad  \textsuperscript{2}Chung-Ang University \\[1ex]
  \texttt{\{dongbin413,jinseong,jaewook\}@snu.ac.kr, hokikim@cau.ac.kr}
}

\begin{document}

\maketitle

\renewcommand{\thefootnote}{\fnsymbol{footnote}}
\footnotetext[1]{Corresponding authors}

\begin{abstract}

Time series forecasting is crucial for applications across multiple domains and various scenarios. Although Transformers have dramatically advanced the landscape of forecasting, their effectiveness remains debated. Recent findings have indicated that simpler linear models might outperform complex Transformer-based approaches, highlighting the potential for more streamlined architectures. In this paper, we shift the focus from evaluating the overall Transformer architecture to specifically examining the effectiveness of self-attention for time series forecasting. To this end, we introduce a new architecture, Cross-Attention-only Time Series transformer (CATS), that rethinks the traditional transformer framework by eliminating self-attention and leveraging cross-attention mechanisms instead. 
By establishing future horizon-dependent parameters as queries and enhanced parameter sharing, our model not only improves long-term forecasting accuracy but also reduces the number of parameters and memory usage. Extensive experiment across various datasets demonstrates that our model achieves superior performance with the lowest mean squared error and uses fewer parameters compared to existing models.
The implementation of our model is available at: \color{blue}{\href{https://github.com/dongbeank/CATS}{https://github.com/dongbeank/CATS}}.
\end{abstract}

\section{Introduction}

Time series forecasting plays a critical role within the machine learning society, given its applications ranging from financial forecasting to medical diagnostics. To improve the accuracy of predictions, researchers have extensively explored and developed various models. These range from traditional statistical methods to modern deep learning techniques. Most notably, Transformer \cite{vaswani2017attention} has brought about a paradigm shift in time series forecasting, resulting in numerous high-performance models, such as Informer \cite{zhou2021informer}, Autoformer \cite{wu2021autoformer}, Pyraformer \cite{liu2021pyraformer}, FEDformer \cite{zhou2022fedformer}, and Crossformer \cite{zhang2022crossformer}.
This line of work establishes new benchmarks for high performance in time series forecasting.

However, \citet{zeng2023transformers} have raised questions about the effectiveness of Transformer-based time series forecasting models especially for long term time series forecasting. Specifically, their experiments demonstrated that simple linear models could outperform these Transformer-based approaches, thereby opening new avenues for research into simpler architectural frameworks.  Indeed, the following studies \cite{liu2022non, ekambaram2023tsmixer} have further validated that these linear models can be enhanced by incorporating additional features.

Despite these developments, the effectiveness of each components in Transformer architecture in time series forecasting remains a subject of debate. \citet{nie2023patchtst} introduced an encoder-only Transformer that utilizes patching rather than point-wise input tokens, which exhibited improved performance compared to linear models. \citet{zeng2023transformers} also highlighted potential shortcomings in simpler linear networks, such as their inability to capture temporal dynamics at change points \cite{van2020evaluation} compared to Transformers.
Consequently, while a streamlined architecture may be beneficial, it is imperative to critically evaluate which elements of the Transformer are necessary and which are not for time series modeling.

In light of these considerations, our study shifts focus from the overall architecture of the Transformer to a more specific question: \textbf{Are self-attentions effective for time series forecasting?} While this question is also noted in \cite{zeng2023transformers}, their analysis was limited to substituting attention layers with linear layers, leaving substantial room for potential model design when focusing on Transformers. Furthermore, the issue of \textit{temporal information loss} (i.e., permutation-invariant and anti-order characteristics of self-attention) is predominantly caused by the use of self-attention rather than the Transformer architecture itself.
Therefore, we aim to resolve the issues of self-attention and therefore propose a new forecasting architecture that achieves higher performance with a more efficient structure.

\begin{figure*}[t!]
  \centering
  \includegraphics[width=1.\linewidth]{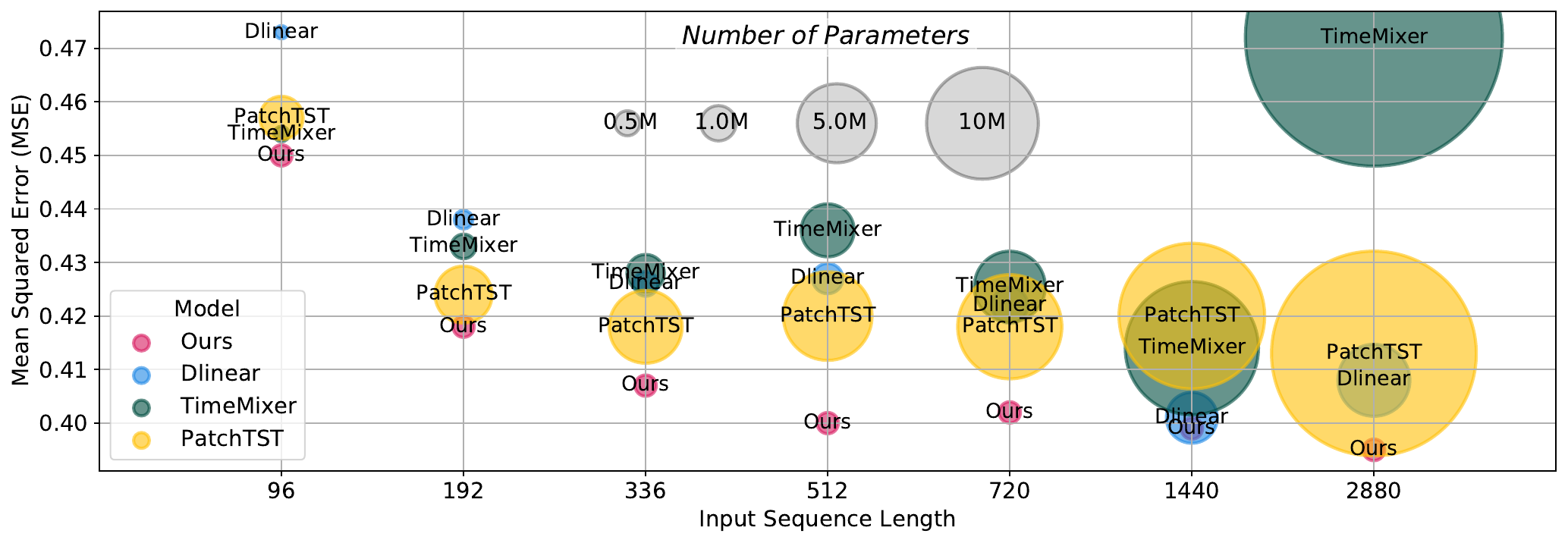}
  \caption{Experimental results illustrating the mean squared error (MSE) and the number of parameters with varying input sequence lengths on ETTm1. Each bubble represents a different model, with the bubble size indicating the number of parameters in millions—larger bubbles denote models with more parameters. Our model consistently shows the lowest MSE (i.e., best performance) with fewer parameters even for longer input sequences. The detailed results can be found in \Tabref{table:ettm1_param}.}
  \label{fig:vis}
\end{figure*}

In this paper, we introduce a novel forecasting architecture named Cross-Attention-only Time Series transformer (CATS) that simplifies the original Transformer architecture by eliminating all self-attentions and focusing on the potential of cross-attentions. Specifically, our model establishes future horizon-dependent parameters as queries and treats past time series data as key and value pairs. This allows us to enhance parameter sharing and improve long-term forecasting performance. As shown in Figure \ref{fig:vis}, our model shows the lowest mean squared error (i.e., better forecasting performance) even for longer input sequences and with fewer parameters than existing models. Moreover, we demonstrate that this simplified architecture can provide a clearer understanding of how future predictions are derived from past data with individual attention maps for the specific forecasting horizon. Finally, through extensive experiments, we show that our proposed model not only achieves state-of-the-art performance but also requires fewer parameters and less memory consumption compared to previous Transformer models across various time series datasets.

\section{Related Work}

\paragraph{Time Series Transformers}
Transformer models \cite{vaswani2017attention} have shown effective
in various domains \cite{dosovitskiy2021an,devlin2019bert,radford2018improving}, with a novel encoder-decoder structure with self-attention, masked self-attention, and cross-attention. The self-attention mechanism is a key component for extracting semantic correlations between paired elements, even with identical input elements; however, autoregressive inference with self-attention requires quadratic time and memory complexity. Therefore, Informer \cite{zhou2021informer} proposed directly predicting multi-steps, and a line of work, such as Autoformer \cite{wu2021autoformer}, FEDformer \cite{zhou2022fedformer}, and Pyraformer \cite{liu2021pyraformer}, investigated the complexity issue in time series transformers. Simultaneously, unique properties of time series, such as stationarity \cite{liu2022non}, decomposition \cite{wu2021autoformer}, frequency features \cite{zhou2022fedformer}, or cross-dimensional properties \cite{zhang2022crossformer} were employed to modify the attention layer for forecasting tasks.
Recently, researchers have investigated the essential architecture in Transformers to capture long-term dependencies. PatchTST \cite{nie2023patchtst} became a de-facto standard Transformer model by patching the time series input in a channel-independence manner, which is widely used in following Transformer-based forecasting models \cite{liu2024itransformer, goswami2024moment}.
On the other hand, \citet{das2023decoder} emphasized the importance of decoder-only forecasting models, while they focused on zero-shot using pre-trained language models. However, none of them have investigated the importance of cross-attention for time series forecasting. 

\paragraph{Temporal Information Encoding}
Fixed temporal order in time series is the distinct property of time series, in contrast to the language domain where semantic information does not heavily depend on the word ordering \cite{devlin2019bert}. Thus, some researchers have used learnable positional encoding in Transformers to embed time-dependent properties  \cite{li2019enhancing,wu2021autoformer}. 
However, \citet{zeng2023transformers} first argued that self-attention is not suitable for time series due to its permutation invariant and anti-order properties. While they focus on building complex representations, they are inefficient in maintaining the original context of historical and future values. They rather proposed linear models without any embedding layer and demonstrated that it can achieve better performance than Transformer models, particularly showing robust performance to long input sequences. 
Recent linear time series models outperformed previous Transformer models with simple architectures by focusing on pre-processing and frequency-based properties \cite{li2023revisiting, chen2023tsmixer, wang2024timemixer}.
On the other hand, \citet{woo2023learning} investigated the new line of works of time-index models, which try to model the underlying dynamics with given time stamps.
These related works imply that preserving the order of time series sequences plays a crucial role in time series forecasting.

\section{Revisiting Self-Attention in Time Series Forecasting} \label{sec:revisiting}
\paragraph{Motivation of Self-Attention Removal}

\begin{figure}[ht!]
    \centering
    \subfloat[Original PatchTST \label{fig:overlapping}]{%
       \includegraphics[width=0.3\linewidth]{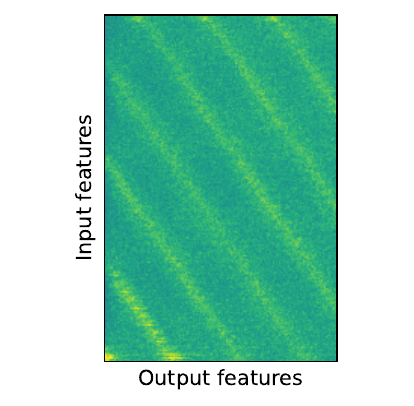}}
    \subfloat[PatchTST w/ non-overlapping \label{fig:non_overlapping}]{%
       \includegraphics[width=0.3\linewidth]{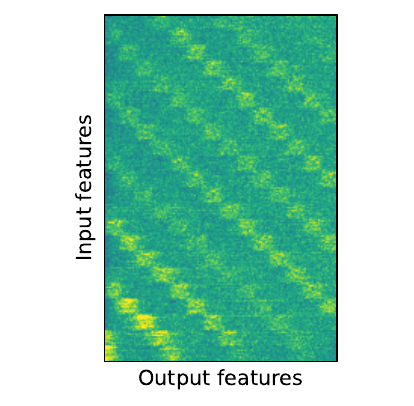}}
    \subfloat[PatchTST w/o self-attn \label{fig:satolinear}]{%
       \includegraphics[width=0.3\linewidth]{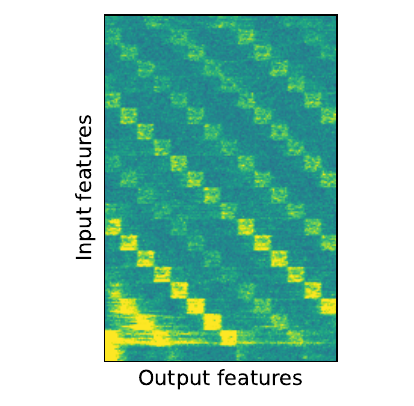}}
    \caption{Absolute values of weights in the final linear layer for different PatchTST variations. The distinct patterns reveal how each model captures temporal information.}
    \label{fig:linear_weights}
\end{figure}

Following the concerns about the effectiveness of self-attention on temporal information preservation \citep{zeng2023transformers}, we conduct an experiment using PatchTST \citep{nie2023patchtst}. We consider three variations of the PatchTST model: the original PatchTST with overlapping patches with length 16 and stride 8 (\Figref{fig:overlapping}); a modified PatchTST with non-overlapping patches with length 24 (\Figref{fig:non_overlapping}); and a version where self-attention is replaced by a linear embedding layer, using non-overlapping patches with length 24 (\Figref{fig:satolinear}). This setup allows us to isolate the effects of self-attention on temporal information preservation, while controlling for the impact of patch overlap.

\Figref{fig:linear_weights} illustrates the absolute values of the weights in the final linear layer for these model variations. Compared to the original PatchTST (\Figref{fig:overlapping}), both non-overlapping versions (\Figref{fig:non_overlapping} and \Figref{fig:satolinear}) show more vivid patterns. The version with linear embedding (\Figref{fig:satolinear}) demonstrates the clearest capture of temporal information, suggesting that the self-attention mechanism itself may not be necessary for capturing temporal information.

In \Tabref{table:self_attention_impact}, we summarize the forecasting performance of the original PatchTST (\Figref{fig:overlapping}) and PatchTST without self-attention (\Figref{fig:satolinear}). PatchTST without self-attention consistently improves or maintains performance across all forecasting horizons. Specifically, the original version with self-attention shows lower performance for longer forecast horizons. This result suggests that self-attention may not only be unnecessary for effective time series forecasting but could even hinder
\begin{wrapfigure}{r}{0.33\textwidth}
\vspace{-0cm}
\centering
\captionof{table}{Effect of self-attention in PatchTST on forecasting performance (MSE) on ETTm1.}
\label{table:self_attention_impact}
\resizebox{0.33\columnwidth}{!}{%
\begin{tabular}{c|c|c}
\toprule
\textbf{Horizon} & \textbf{original} & \textbf{w/o self-attn} \\ \midrule
96  & \textbf{0.290} & \textbf{0.290} \\
192 & 0.332 & \textbf{0.328} \\
336 & 0.366 & \textbf{0.359} \\
720 & 0.416 & \textbf{0.414} \\
\bottomrule
\end{tabular}%
}
\vspace{-0.3cm}
\end{wrapfigure}
performance in certain cases. Therefore, better performance of w/o self-attention challenges the conventional belief regarding the importance of self-attention mechanisms in Transformer-based models for time series forecasting tasks.

Our findings offer new insights into the role of self-attention in time series forecasting. As shown in \Figref{fig:linear_weights} and \Tabref{table:self_attention_impact}, replacing self-attention with a linear layer not only captures clear temporal patterns but also results in significant performance improvements, particularly for longer forecast horizons. These results highlight potential areas for enhancing the handling of temporal information, beyond addressing the well-known concerns regarding computational complexity.

\paragraph{Rethinking Transformer Design}

\begin{figure}[ht!]
    \centering
    \subfloat[{Transformer} \label{fig:transformer}]{%
       \includegraphics[width=0.25965\linewidth]{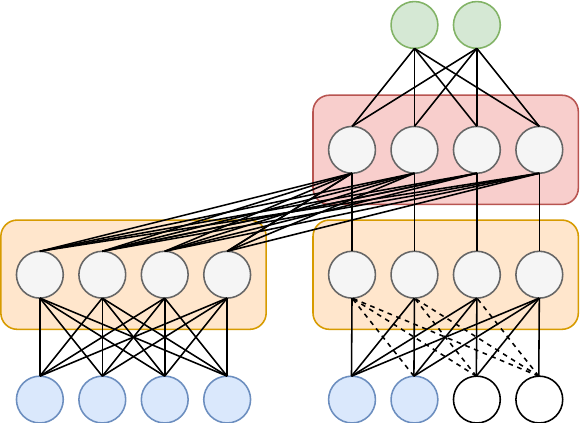}}
     \quad \quad 
    \subfloat[{Encoder} \label{fig:encoder}]{%
       \includegraphics[width=0.1193\linewidth]{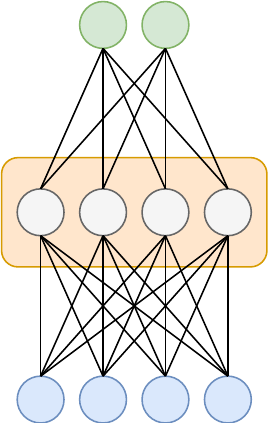}}
     \quad \quad 
    \subfloat[{Linear model} \label{fig:dlinear}]{%
       \includegraphics[width=0.21754\linewidth]{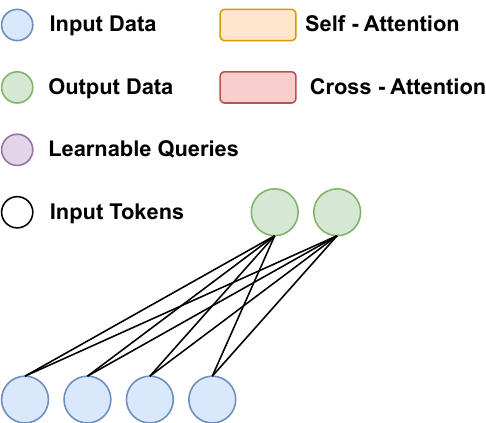}}
    \subfloat[{Ours} \label{fig:cats}]{%
       \includegraphics[width=0.20351\linewidth]{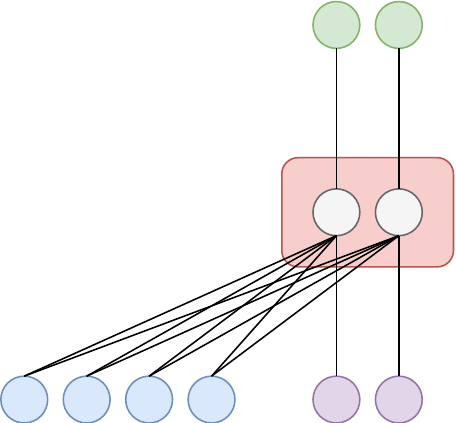}}
    \caption{Illustration of existing time series forecasting architectures and the proposed architecture.}
    \label{fig:model_architectures}
\end{figure}

Given the challenges associated with self-attention in time series forecasting, we propose a fundamental rethinking of the Transformer architecture for this task. \Figref{fig:model_architectures} illustrates the differences between existing architectures and our proposed approach. Traditional Transformer architectures (\Figref{fig:transformer}) and encoder-only models (\Figref{fig:encoder}) rely heavily on self-attention mechanisms, which may lead to temporal information loss. In contrast, \citet{zeng2023transformers} proposed a simplified linear model, DLinear (\Figref{fig:dlinear}), which removes all Transformer-based components. While this approach reduces computational load and potentially avoids some temporal information loss, it may struggle to capture complex temporal dependencies.

To address these challenges while preserving the advantages of Transformer architectures, we propose the Cross-Attention-only Time Series transformer (CATS), depicted in \Figref{fig:cats}. Our approach removes all self-attention layers and focuses solely on cross-attention, aiming to better capture temporal dependencies while maintaining the structural advantages of the transformer architecture. In the following section, we will introduce our CATS model in detail, explaining our key innovations including a novel use of cross-attention, efficient parameter sharing, and adaptive masking techniques.

\section{Proposed Methodology}\label{sec:proposed}

\subsection{Problem Definition and Notations} \label{sec:proposed:problem_definition}

A multivariate time series forecasting task aims to predict future values $\boldsymbol{\tilde{X}}=\{\rvx_{L+1},\ldots,\rvx_{L+T}\} \in \mathbb{R}^{M\times T}$ with the prediction $\boldsymbol{\hat{X}}=\{\hat{\rvx}_{L+1},\ldots,\hat{\rvx}_{L+T}\} \in \mathbb{R}^{M\times T}$ based on past datasets $\boldsymbol{X}=\{\rvx_1,\ldots,\rvx_L\} \in \mathbb{R}^{M\times L}$. Here, $T$ represents the forecasting horizon, $L$ denotes the input sequence length, and $M$ represents the dimension of time series data.

In traditional time series transformers, we feed the historical multivariate time series $\boldsymbol{X}$ to embedding layers, resulting in the historical embedding $\boldsymbol{H} \in \mathbb{R}^{D\times L}$. Here, $D$ is the embedding size. Note that, in channel-independence manners, the multivariate input is considered to separate univariate time series $\rvx \in \mathbb{R}^{1\times L}$. With patching \citep{nie2023patchtst}, univariate time series $\rvx$ transforms into patches $\rvp = \texttt{Patch}(\rvx)\in \mathbb{R}^{P\times N_L}$ where $P$ is the size of each patch and $N_L$ is the number of input patches. Similar to non-patching situations, patches are fed to embedding layers $\boldsymbol{P} = \texttt{Embedding}(\rvp)\in \mathbb{R}^{D\times N_L}$.

\begin{figure*}[t!]
  \centering
  \includegraphics[width=1.\linewidth]{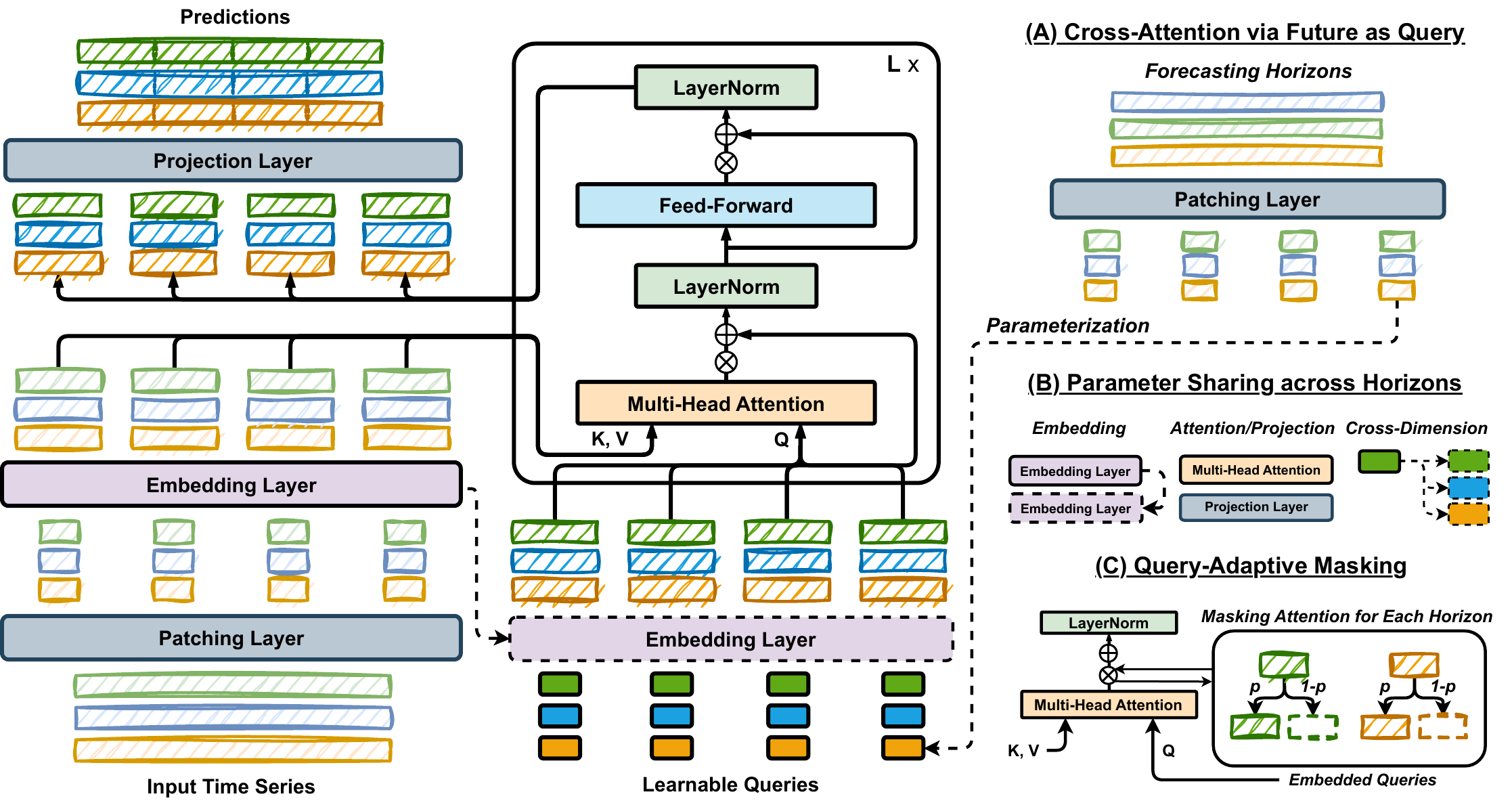}
  \caption{Illustration on the proposed model architecture. Our model removes all self-attentions from the original Transformer structure and focuses on cross-attentions. To fully utilize the cross-attention, we conceptualize the future horizon as queries and use the input time series (i.e., past time series) as keys and values (Fig. A). This simplified structure enables us to enhance the parameter sharing across forecasting horizons (Fig. B) and make use of query-adaptive masking (Fig. C) for performance.}
  \label{fig:model}
\end{figure*}

\subsection{Model Structure} \label{sec:proposed:structure}

Our proposed architecture consists of three key components: (A) \textit{Cross-Attention with Future as Query}, (B) \textit{Parameter Sharing across Horizons}, and (C) \textit{Query-Adaptive Masking}. \Figref{fig:model} illustrates how our model modifies the traditional transformer structure.

By focusing solely on cross-attention, our approach allows us to maintain the periodic properties of time series, which self-attention, with its permutation-invariant and anti-order characteristics, struggles to capture. Furthermore, we leverage advanced architectural designs of time series transformers, such as patching \citep{nie2023patchtst}, which linear models cannot utilize.
The following subsections provide detailed descriptions of each component of our CATS model, explaining how these elements work together to address the challenges of time series forecasting.

\paragraph{Cross-Attention via Future as Query}

Similar to self-attention, the cross-attention mechanism employs three elements: key, query, and value. The distinctive feature of cross-attention is that the query originates from a different source than the key or value. Generally, the query component aims to identify the most relevant information among the keys and uses it to extract crucial data from the values \cite{chen2021crossvit, zhang2023fcaformer}. In the realm of time series forecasting, where predictions are often made for a specific target horizon—such as forecasting 10 steps ahead. Therefore, within this concept of forecasting, we argue that each future horizon should be regarded as a question, i.e., an independent query.

To implement this, we establish horizon-dependent parameters as learnable queries. As shown in \Figref{fig:model}, we begin by creating parameters for the specified forecasting horizon. For each of these virtualized parameters, we assign a fixed number of parameters to represent the corresponding horizon as learnable queries $\rvq$. For example, $\rvq_{i}$ is a horizon-dependent query at $L+i$. When patching is applied, these queries are then processed independently; each learnable query $\rvq\in\mathbb{R}^P$ is first fed into the embedding layer, and then fed into the multi-head attention with the embedded input time series patches as the key and value. 

Based on these new query parameters, we can utilize a cross-attention-only structure in the decoder, resulting in an advantage in efficiency. In \Tabref{tab:complexity_comparison}, we summarize the time complexity of recent transformer models and ours. The results indicate that our method only requires the time complexity of $\mathcal{O}({LT}/{P^2})$, where most of the transformer-based models require $\mathcal{O}({L}^2)$ except FEDformer and Pyraformer. However, since these two models have an encoder-decoder and a relatively huge amount of parameters, they require 10x and 4x computational times than ours, respectively.

\begin{table}[t!]
    \centering
    \caption{Time complexity of transformer-based models to calculate attention outputs. Time refers to the inference time obtained by averaging 10 runs under $L=96$ and $T=720$ on Electricity.}
\resizebox{1\columnwidth}{!}{%
    \begin{tabular}{lccr|lccr}
        \toprule
        \textbf{Method} & \textbf{Encoder} & \textbf{Decoder} & \multicolumn{1}{c}{\textbf{Time}} & \textbf{Method} & \textbf{Encoder} & \textbf{Decoder} & \multicolumn{1}{c}{\textbf{Time}} \\
        \midrule
        Transformer \cite{liu2024itransformer} & $\mathcal{O}(L^2)$ & $\mathcal{O}(T(T + L))$ & 10.4ms &
        Informer \cite{zhou2021informer} & $\mathcal{O}(L \log L)$ & $\mathcal{O}(T(T + \log L))$ & 13.5ms\\
        Autoformer \cite{wu2021autoformer} & $\mathcal{O}(L \log L)$ & $\mathcal{O}\left(\left(L/2 + H\right)\log\left(L/2 + T\right)\right)$ & 24.1ms &
        Pyraformer \cite{liu2021pyraformer} & $\mathcal{O}(L)$ & $\mathcal{O}(T(T + L))$ & 11.2ms\\
        FEDformer \cite{zhou2022fedformer} & $\mathcal{O}(L)$ & $\mathcal{O}\left(L/2 + H\right)$ & 69.3ms &
        Crossformer \cite{zhang2022crossformer} & $\mathcal{O}\left({M}L^2/{P^2}\right)$ & $\mathcal{O}\left({M}T(T + L)/{P^2}\right)$ & 30.6ms\\
        PatchTST \cite{nie2023patchtst} & $\mathcal{O}\left({L^2}/{P^2}\right)$ & - & 7.6ms&
        CATS (Ours) & - & $\mathcal{O}\left({LT}/{P^2}\right)$&7.0ms \\
        \bottomrule
    \end{tabular}%
    }
    \label{tab:complexity_comparison}
\end{table}

\paragraph{Parameter Sharing across Horizons}

One of the strongest benefits of cross-attention via future horizon as a query $\rvq$ is that each cross-attention is only calculated on the values from a single forecasting horizon and the input time series. Mathematically, for a prediction of future value $\hat{\rvx}_{L+i}$ can be expressed as a function solely dependent on the past samples $\boldsymbol{X}=[\rvx_1, ..., \rvx_L$] and $\rvq_{i}$, independent of $\rvq_{j}$ for all $i\neq j$ or $i$ and $j$ are not in the same patch.

This independent forecasting mechanism offers a notable advantage; a higher level of parameter sharing. As demonstrated in \cite{nie2023patchtst}, significant reductions in the required number of parameters can be achieved in time series forecasting through parameter sharing between inputs or patches, enhancing computational efficiency. Regarding this, we propose parameter sharing across all possible layers — the embedding layer, multi-head attention, and projection layer — for every horizon-dependent query $\rvq$. In other words, all horizon queries $\rvq_{1}, \dots, \rvq_{T}$ or $\rvq_{1}, \dots, \rvq_{{N_T}}$ share the same embedding layer used for the input time series $\rvx_1, \dots, \rvx_L$  or patches $\rvp_1, \dots, \rvp_{N_L}$ before proceeding to the cross-attention layer, respectively. Furthermore, to maximize the parameter sharing, we also propose cross-dimension sharing that uses the same query parameters for all dimensions.

For the multi-head attention and projection layers, we apply the same algorithm across horizons. Notably, unlike the approach in PatchTST \cite{nie2023patchtst}, we also share the projection layer for each prediction. Specifically, PatchTST, being an encoder-only model, employs a fully connected layer as the projection layer for the encoder outputs $\boldsymbol{P} \in \mathbb{R}^{D \times N_L}$, resulting in $(D\times N_L) \times T$ parameters. In contrast, our model first processes raw queries $\rvq = [\rvq_1, \ldots, \rvq_{N_T}] \in \mathbb{R}^{P \times N_T}$. These queries are then embedded through the cross-attention mechanism, resulting in $\boldsymbol{Q} = [\boldsymbol{q}_1, \ldots, \boldsymbol{q}_{N_T}] \in \mathbb{R}^{D \times N_T}$. The final projection uses shared parameters $W \in \mathbb{R}^{P \times D}$, producing an output $W\boldsymbol{Q} \in \mathbb{R}^{P \times N_T}$. Thus, our number of parameters for this projection becomes $P \times D$, which is not proportionally increasing to $T$. This approach significantly reduces time complexity during both the training and inference phases.

\begin{wrapfigure}{r}{0.4\textwidth}
\vspace{-0.43cm}
\centering
\captionof{table}{Effect of parameter sharing across horizons on the number of parameters for different forecasting horizons on ETTh1.}
\label{table:parameter_sharing}
\resizebox{0.37\columnwidth}{!}{%
\begin{tabular}{c|r|r}
\toprule
\textbf{Horizon} & \textbf{w/ sharing} & \textbf{w/o sharing} \\ \midrule
96  & 355,320 & 404,672 \\
192 & 355,416 & 552,320 \\
336 & 355,560 & 958,112 \\
720 & 355,944 & 3,121,568 \\
\bottomrule
\end{tabular}%
}
\vspace{-0.45cm}
\end{wrapfigure}

In \Tabref{table:parameter_sharing}, we outline the impact of parameter sharing across different forecasting horizons. In contrast to the model without parameter sharing, which exhibits a rapid increase in parameters as the forecasting horizon extends, our model, which shares all layers including the projection layer, maintains a nearly consistent number of parameters. 

Additionally, all operations, including embedding and multi-head attention, are performed independently for each learnable query. This implies that the forecast for a specific horizon does not depend on other horizons. Such an approach allows us to generate distinct attention maps for each forecasting horizon, providing a clear understanding of how each prediction is derived. Please refer to \Secref{sec:toy}.

\paragraph{Query-Adaptive Masking}

Parameter sharing across horizons enhances the efficiency of our proposed architecture and simplifies the model. However, we observed that a high degree of parameter sharing could lead to overfitting to the keys and values (i.e., past time series data), rather than the queries (i.e., forecasting horizon). Specifically, the model may converge to generate similar or identical predictions, $\hat{\rvx}_{L+i}$ and $\hat{\rvx}_{L+j}$, despite receiving different horizon queries,  $\rvq_{i}$ and $\rvq_{j}$ (i.e., the target horizons differ).

Therefore, to ensure the model focuses on each horizon-dependent query $\rvq$, we introduce a new technique that masks the attention outputs. As illustrated in the right-bottom figure of \Figref{fig:model}, for each horizon, we apply a mask to the direct connection from Multi-Head Attention to LayerNorm with a probability $p$. This mask prevents access to the input time series information, resulting in only the query to influence future value predictions. This selective disconnection, rather than the application of dropout in the residual connections, helps the layers to concentrate more effectively on the forecasting queries. We note that this approach can be related to stochastic depth in residual networks \cite{huang2016deep}. The stochastic depth technique has proven effective across various tasks, such as vision tasks \cite{strudel2021segmenter, yang2022lite}. To the best of our knowledge, this is the first application of stochastic depth in Transformers for time series forecasting. A detailed analysis of query-adaptive masking can be found in Appendix.

In summary, the framework described in this section, including cross-attention via future as query, parameter sharing across horizons, and query-adaptive masking, is named the \textbf{Cross-Attention-only Time Series transformer (CATS).}

\section{Experiments}

In this section, we provide extensive experiments to provide the benefits of our proposed framework, CATS, including forecasting performance and computational efficiency. To this end, we use 7 different real-world datasets and 9 baseline models.
For datasets, we use Electricity, ETT (ETTh1, ETTh2, ETTm1, and ETTm2), Weather, Traffic, and M4. These datasets are provided in \cite{wu2021autoformer} and \cite{wu2022timesnet} for time series forecasting benchmark, detailed in Appendix.

For baselines, we utilize a wide range of various baselines, including the state-of-the-art long-term time series forecasting model TimeMixer \citep{wang2024timemixer}, PatchTST \citep{nie2023patchtst}, Timesnet \citep{wu2022timesnet}, Crossformer \citep{zhang2022crossformer}, MICN \citep{wang2022micn}, FiLM \citep{zhou2022film}, DLinear \citep{zeng2023transformers}, Autoformer \citep{wu2021autoformer}, and Informer \citep{zhou2021informer}. For both long-term and short-term time series forecasting results, we report performance of our model alongside the results of other models as presented in TimeMixer \citep{wang2024timemixer}, ensuring a consistent comparison across all baselines. We used 4 NVIDIA RTX 4090 24GB GPUs with 2 Intel(R) Xeon(R) Gold 5218R CPUs @ 2.10GHz for all experiments.

\subsection{Long-Term Time Series Forecasting Results} 

To ease comparison, we follow the settings of \cite{wang2024timemixer} for long-term forecasting, using various forecast horizons with a fixed 96 input sequence length. Detailed settings are provided in the Appendix. \Tabref{table:full_result_96} summarizes the forecasting performance across all datasets and baselines. Our proposed model, CATS, demonstrates superior performance in multivariate long-term forecasting tasks across multiple datasets. CATS consistently achieves the lowest Mean Squared Error (MSE) and Mean Absolute Error (MAE) on the Traffic dataset for all forecast horizons, outperforming all other models. For the Weather, Electricity, and ETT datasets, CATS shows competitive performance, achieving the best results on most forecast horizons. This indicates that CATS effectively captures underlying patterns in diverse time series data, highlighting its capability to handle complex temporal dependencies. Additional experiments with longer input sequence lengths of 512 are provided in the Appendix.

\begin{table}[ht!]
\setlength{\tabcolsep}{3pt}
\centering
\caption{Multivariate long-term forecasting results with recent forecasting models and ours for unified hyperparameter settings. The best results are in \textbf{bold} and the second best are \underline{underlined}. Full results are provided in Appendix.}
\label{table:result_96}
\resizebox{1\columnwidth}{!}{%
\begin{tabular}{c|c|cc|cc|cc|cc|cc|cc|cc|cc|cc|cc}
\toprule [1.2pt]
\multicolumn{2}{c}{Models}
&\multicolumn{2}{c}{CATS}&\multicolumn{2}{c}{TimeMixer}&\multicolumn{2}{c}{PatchTST}&\multicolumn{2}{c}{Timesnet}&\multicolumn{2}{c}{Crossformer} & \multicolumn{2}{c}{MICN} & \multicolumn{2}{c}{FiLM}&\multicolumn{2}{c}{DLinear} & \multicolumn{2}{c}{Autoformer}& \multicolumn{2}{c}{Informer}\\ 
\cmidrule(lr){3-4}\cmidrule(lr){5-6}\cmidrule(lr){7-8}\cmidrule(lr){9-10}\cmidrule(lr){11-12}\cmidrule(lr){13-14}\cmidrule(lr){15-16}\cmidrule(lr){17-18}\cmidrule(lr){19-20}\cmidrule(lr){21-22}

\multicolumn{2}{c}{Metric}  & MSE & MAE & MSE & MAE & MSE & MAE & MSE & MAE & MSE & MAE & MSE & MAE & MSE & MAE & MSE & MAE & MSE & MAE & MSE & MAE \\ \midrule

\multirow{4}{*}{\begin{turn}{90}Weather\end{turn}}

 & 96  & \textbf{0.161} & \textbf{0.207} & \underline{0.163} & \underline{0.209} & 0.186 & 0.227 & 0.172 & 0.220 & 0.195 & 0.271 & 0.198 & 0.261 & 0.195 & 0.236 & 0.195 & 0.252  & 0.266 & 0.336 & 0.300 & 0.384 \\
 & 192 & \textbf{0.208} & \textbf{0.250} & \textbf{0.208} & \textbf{0.250} & 0.234 & 0.265 & 0.219 & \underline{0.261} & \underline{0.209} & 0.277 & 0.239 & 0.299 & 0.239 & 0.271 & 0.237 & 0.295  & 0.307 & 0.367 & 0.598 & 0.544 \\
 & 336 & 0.264 & \underline{0.290} & \underline{0.251} & \textbf{0.287} & 0.284 & 0.301 & \textbf{0.246} & 0.337 & 0.273 & 0.332 & 0.285 & 0.336 & 0.289 & 0.306 & 0.282 & 0.331  & 0.359 & 0.395 & 0.578 & 0.523 \\
 & 720 & \underline{0.342} & \textbf{0.341} & \textbf{0.339} & \textbf{0.341} & 0.356 & \underline{0.349} & 0.365 & 0.359 & 0.379 & 0.401 & 0.351 & 0.388 & 0.361 & 0.351 & 0.345 & 0.382  & 0.419 & 0.428 & 1.059 & 0.741 \\ \midrule

\multirow{4}{*}{\begin{turn}{90}Electricity\end{turn}}
 & 96  & \textbf{0.149} & \textbf{0.237} & \underline{0.153} & \underline{0.247} & 0.190 & 0.296 & 0.168 & 0.272 & 0.219 & 0.314 & 0.180 & 0.293 & 0.198 & 0.274 & 0.210 & 0.302  & 0.201 & 0.317 & 0.274 & 0.368 \\
 & 192 & \textbf{0.163} & \textbf{0.250} & \underline{0.166} & \underline{0.256} & 0.199 & 0.304 & 0.184 & 0.322 & 0.231 & 0.322 & 0.189 & 0.302 & 0.198 & 0.278 & 0.210 & 0.305  & 0.222 & 0.334 & 0.296 & 0.386 \\
 & 336 & \textbf{0.180} & \textbf{0.268} & \underline{0.185} & \underline{0.277} & 0.217 & 0.319 & 0.198 & 0.300 & 0.246 & 0.337 & 0.198 & 0.312 & 0.217 & 0.300 & 0.223 & 0.319  & 0.231 & 0.443 & 0.300 & 0.394 \\
 & 720 & \underline{0.219} & \textbf{0.302} & 0.225 & \underline{0.310} & 0.258 & 0.352 & 0.220 & 0.320 & 0.280 & 0.363 & \textbf{0.217} & 0.330 & 0.278 & 0.356 & 0.258 & 0.350  & 0.254 & 0.361 & 0.373 & 0.439 \\ \midrule

\multirow{4}{*}{\begin{turn}{90}Traffic\end{turn}}
 & 96  & \textbf{0.421} & \textbf{0.270} & \underline{0.462} & \underline{0.285} & 0.526 & 0.347 & 0.593 & 0.321 & 0.644 & 0.429 & 0.577 & 0.350 & 0.647 & 0.384 & 0.650 & 0.396  & 0.613 & 0.388 & 0.719 & 0.391 \\
 & 192 & \textbf{0.436} & \textbf{0.275} & \underline{0.473} & \underline{0.296} & 0.522 & 0.332 & 0.617 & 0.336 & 0.665 & 0.431 & 0.589 & 0.356 & 0.600 & 0.361 & 0.598 & 0.370  & 0.616 & 0.382 & 0.696 & 0.379 \\
 & 336 & \textbf{0.453} & \textbf{0.284} & \underline{0.498} & \underline{0.296} & 0.517 & 0.334 & 0.629 & 0.336 & 0.674 & 0.420 & 0.594 & 0.358 & 0.610 & 0.367 & 0.605 & 0.373  & 0.622 & 0.337 & 0.777 & 0.420 \\
 & 720 & \textbf{0.484} & \textbf{0.303} & \underline{0.506} & \underline{0.313} & 0.552 & 0.352 & 0.640 & 0.350 & 0.683 & 0.424 & 0.613 & 0.361 & 0.691 & 0.425 & 0.645 & 0.394  & 0.660 & 0.408 & 0.864 & 0.472 \\ \midrule

\multirow{4}{*}{\begin{turn}{90}ETT (Avg)\end{turn}}
&96  & \textbf{0.289} & \textbf{0.339} & \underline{0.290} & \textbf{0.339} & 0.326 & 0.362 & 0.312 & \underline{0.355} & 0.465 & 0.456 & 0.340 & 0.388 & 0.324 & 0.358 & 0.319 & 0.368 & 0.389 & 0.415 & 1.414 & 0.816 \\
&192 & \textbf{0.348} & \underline{0.374} & \underline{0.350} & \textbf{0.373} & 0.388 & 0.397 & 0.365 & 0.385 & 0.553 & 0.518 & 0.408 & 0.431 & 0.384 & 0.393 & 0.399 & 0.418 & 0.448 & 0.443 & 1.985 & 0.989 \\
&336 & \textbf{0.376} & \textbf{0.395} & \underline{0.390} & \underline{0.404} & 0.426 & 0.423 & 0.455 & 0.421 & 0.686 & 0.584 & 0.479 & 0.476 & 0.428 & 0.423 & 0.469 & 0.463 & 0.491 & 0.473 & 2.101 & 1.101 \\
&720 & \textbf{0.434} & \textbf{0.433} & \underline{0.439} & \underline{0.438} & 0.464 & 0.455 & 0.467 & 0.455 & 1.038 & 0.754 & 0.597 & 0.541 & 0.481 & 0.459 & 0.596 & 0.537 & 0.533 & 0.504 & 2.343 & 1.163 \\  \bottomrule[1.2pt]
 
\end{tabular} 
}
\end{table}

\subsection{Efficient and Robust Forecasting for Long Input Sequences}\label{sec:efficient}
\citet{zeng2023transformers} observed that many models experience a decline in performance when using long input sequences for time series forecasting. To address this, some approaches have been developed to capture long-term dependencies. For instance, TimeMixer \cite{wang2024timemixer} employs linear models with mixed scale, and PatchTST \cite{nie2023patchtst} utilizes an encoder network to encode long-term information. However, these models still have computational issues, particularly in terms of escalating memory and parameter requirements. Thus, in this subsection, we provide a comparison between previous models and ours in terms of efficient and robust forecasting for long input sequences.

First of all, to provide a fair comparison, we summarize the number of parameters, GPU memory consumption, and forecasting performance of comparison models with varying input lengths. As summarized in \Tabref{table:ettm1_param}, existing complex models, such as PatchTST and TimeMixer, suffer from increased parameters and computational burdens when performing forecasting with long input lengths. Although DLinear uses fewer parameters and less GPU memory, its performance is limited due to its linear structure in capturing non-linearity patterns. Considering both performance and efficiency, the proposed model demonstrates robust performance improvement even with longer input lengths. In Appendix, we provide additional experimental results supporting these findings.

\begin{table}[t!]
\setlength{\tabcolsep}{3pt}
\centering
\caption{Comparison of models with the number of parameters, GPU memory consumption, and MSE across different input sequence lengths on ETTm1. Full results with more diverse input lengths are provided in Appendix.}
\label{table:ettm1_param}
\resizebox{1\columnwidth}{!}{%
\begin{tabular}{lcrrrcrrrcrrr}
\toprule [1.2pt]
& \multicolumn{4}{c}{\textbf{Parameters}} & \multicolumn{4}{c}{\textbf{GPU Memory}} & \multicolumn{4}{c}{\textbf{MSE}} \\
\cmidrule(lr){2-5}\cmidrule(lr){6-9}\cmidrule(lr){10-13}
\textbf{Input Length} & \multicolumn{1}{c}{336} & \multicolumn{1}{c}{720} & \multicolumn{1}{c}{1440} & \multicolumn{1}{c}{2880} & \multicolumn{1}{c}{336} & \multicolumn{1}{c}{720} & \multicolumn{1}{c}{1440} & \multicolumn{1}{c}{2880}& \multicolumn{1}{c}{336} & \multicolumn{1}{c}{720} & \multicolumn{1}{c}{1440} & \multicolumn{1}{c}{2880} \\ \midrule

PatchTST & 4.3M & 8.7M (2.0x) & 17.0M (4.0x) & 33.6M (7.9x) & 3.5GB & 7.4GB (2.1x) & 22.0GB (6.3x) & 58.6GB (16.9x) & 0.418 & 0.418 & 0.420 & 0.412\\
TimeMixer & 1.1M & 4.1M (3.6x) & 14.2M (12.6x) & 52.9M (46.8x) & 2.9GB & 3.9GB (1.3x) & 5.9GB (2.0x) & 10.3GB (3.6x) & 0.428 & 0.425 & 0.414 & 0.472 \\
DLinear & 0.5M & 1.0M (2.1x) & 2.1M (4.2x) & 4.2M (8.5x) & 1.1GB & 1.1GB (1.0x) & 1.2GB (1.0x) & 1.2GB (1.1x) & 0.426 & 0.422 & 0.401 & 0.408 \\
CATS & 0.4M & 0.4M (1.0x) & 0.4M (1.0x) & 0.4M (1.1x) & 1.9GB & 2.1GB (1.1x) & 2.7GB (1.4x) & 3.8GB (2.0x) & {0.407} & {0.402} & {0.399} & {0.395} \\
\bottomrule[1.2pt]
\end{tabular}
}
\vspace{-0.2cm}
\end{table}

\begin{figure}[ht!]
    \centering
    \subfloat[{Model Performance} \label{fig:model_performance}]{%
       \includegraphics[width=0.255\linewidth]{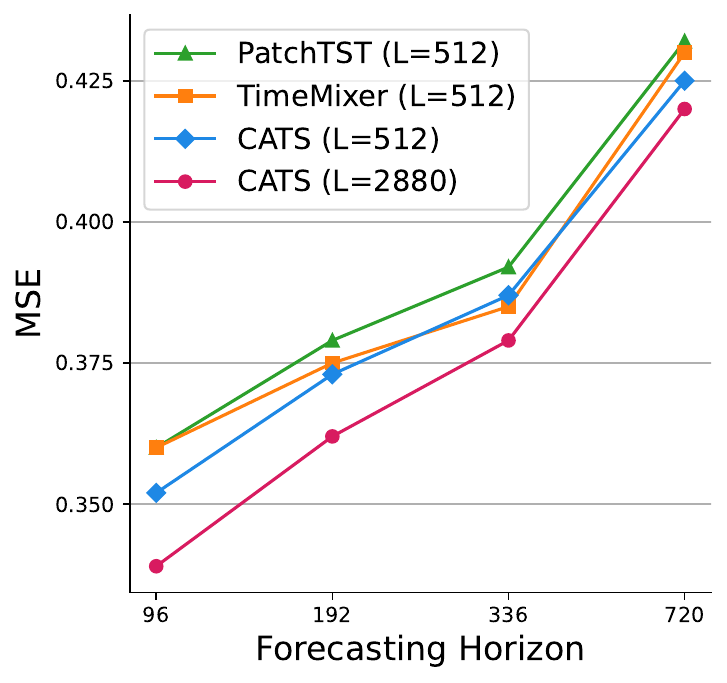}}
    \subfloat[{Parameter Efficiency} \label{fig:parameter_efficiency}]{%
       \includegraphics[width=0.247\linewidth]{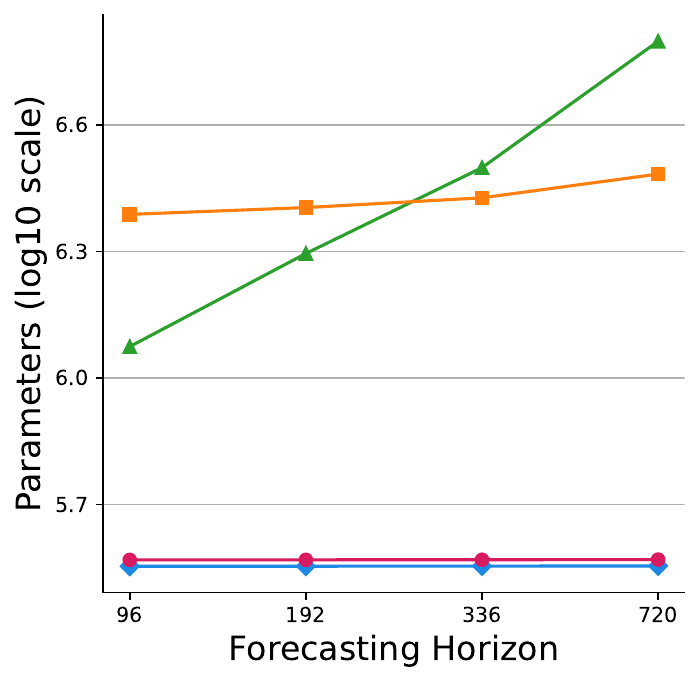}}
    % \quad
    \subfloat[{Memory Efficiency} \label{fig:memory_efficiency}]{%
       \includegraphics[width=0.247\linewidth]{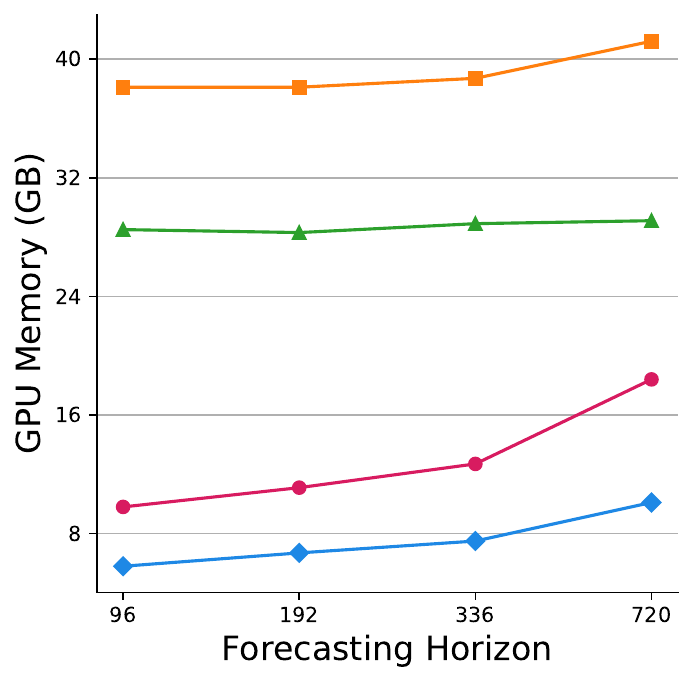}}
    % \quad
    \subfloat[{Running Time Efficiency} \label{fig:run_time_efficiency}]{%
       \includegraphics[width=0.251\linewidth]{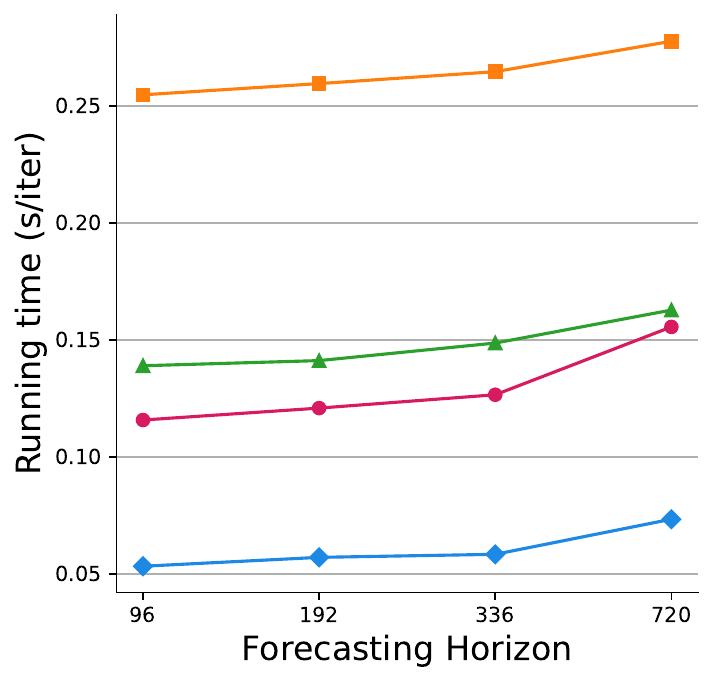}}
    % \quad
    \caption{Efficiency and performance analysis for time series forecasting models. We summarize the forecasting performance, number of parameters, GPU memory consumption, and running time with varying forecasting horizon lengths on Traffic. The running time is averaged from 300 iterations.}
    \label{fig:efficiency_analysis}
\end{figure}

Furthermore, we conduct a deeper comparison between transformer-based models. Especially, TimeMixer \cite{wang2024timemixer} argues that their model outperforms PatchTST \cite{nie2023patchtst} in the setting of long input sequences. Regarding this setting, we also conduct an experiment on $L=512$. We summarize the results in \Figref{fig:efficiency_analysis}. Among these transformer-based models, our model achieves the lowest MSE for most forecasting horizons. Moreover, our model requires even a lower number of parameters, GPU memory, and running time. Especially, for parameter efficiency, CATS shows significant differences even on a log scale due to its efficient parameter-sharing. \Figref{fig:memory_efficiency} highlights GPU memory usage across different forecasting horizons. While PatchTST and TimeMixer consume significantly more memory, CATS maintains a low and stable memory consumption, demonstrating superior memory efficiency. In \Figref{fig:run_time_efficiency} CATS also consistently achieves lower running times compared to PatchTST and TimeMixer. 

Additionally, we also compare the same factors when we use a longer input length $L=2880$. As more input length is used, the forecasting performance of our model outperforms all other models. Most importantly, while the computational complexity increases as input length increases, our model achieves a faster running time, compared to other models with a 512 input sequence length.
Overall, these results emphasize the efficiency and performance advantages of our model, particularly in terms of parameter count, memory usage, and running time.

\subsection{Short-Term Time Series Forecasting Results}

\Tabref{table:Avg_M4_results} summarizes the averaged results for the M4 dataset, comparing the performance of various models. Our CATS model consistently achieved the best results across all metrics. Notably, CATS reduced MASE by 19.94\% compared to PatchTST, a self-attention-only model that suffers from temporal information loss as mentioned in \Secref{sec:revisiting}. CATS, with its cross-attention-only structure, effectively mitigated this issue and captured temporal dependencies more efficiently than previous models. Although TimeMixer, a state-of-the-art linear model, performed well, CATS surpassed it across all metrics. This demonstrates that CATS excels at capturing short-term temporal dependencies, providing superior performance in short-term forecasting tasks.

\begin{table}[ht!]
\vspace{-0.3cm}
\setlength{\tabcolsep}{3pt}
\centering
\caption{Averaged univariate short-term forecasting results in the M4 dataset. The best results are in \textbf{bold} and the second best are \underline{underlined}. Full results are presented in Appendix.}
\label{table:Avg_M4_results}
\resizebox{1\columnwidth}{!}{%
\begin{tabular}{c|c|ccccccccc}
\toprule [1.2pt]
\multicolumn{2}{c}{Models}
&CATS & TimeMixer & Timesnet & PatchTST & MICN & FiLM & DLinear & Autoformer & Informer\\ \midrule

\multirow{3}{*}{\begin{turn}{90}Average\end{turn}}

 & SMAPE & \textbf{11.701} & \underline{11.723} & 11.829 & 13.152 & 19.638 & 14.863 & 13.639 & 12.909 & 14.086 \\
 & MASE  & \textbf{1.557}  & \underline{1.559}  & 1.585  & 1.945  & 5.947  & 2.207  & 2.095  & 1.771  & 2.718  \\
 & OWA   & \textbf{0.838}  & \underline{0.840}  & 0.851  & 0.998  & 2.279  & 1.125  & 1.051  & 0.939  & 1.230  \\ \bottomrule[1.2pt]
\end{tabular} 
}
\vspace{-0.3cm}
\end{table}

\subsection{Replacement of Cross-attention with Self-attention}

In our propose structure, we mainly use cross-attention rather than self-attention due to the forecasting-unfriendly properties of self-attention.
To verify the effectiveness of cross-attention in the proposed structure, we replace the cross-attention layers with self-attention layers while maintaining other structures. In \Tabref{tab:case}, we gradually replace the cross-attention with self-attention (\texttt{SA}) among a total of three cross-attention layers. To maintain the original transformer structure, we set the maximum replacement as two. As shown in this table, we confirm the effectiveness of the cross-attention mechanism compared to using self-attention layers. Specifically, the zero \texttt{SA}, which is our model, shows better performance than using \texttt{SA} for almost all cases except only one case.

\begin{table}[ht!]
\vspace{-0.3cm}
\centering
\caption{Performance comparison on models with three attention layers. We replace one or more cross-attentions (\texttt{CA}) with self-attentions (\texttt{SA}) in our model. In total, there are three cross-attentions in all settngs and `Zero \texttt{SA}' stands for our model. The best results are in \textbf{bold}.}
\label{tab:case}
\resizebox{1\columnwidth}{!}{%
\begin{tabular}{c|cc|cc|cc|cc|cc|cc}
\toprule %\multicolumn{2}{c|}{Model} & \multicolumn{6}{c}{CATS} \\ \midrule
{Dataset} & \multicolumn{6}{c|}{{Electricity}} & \multicolumn{6}{c}{{ETTm1}} \\ \cmidrule(lr){2-7} \cmidrule(lr){8-13}
{Case}  & \multicolumn{2}{c}{{Zero} \texttt{SA}} & \multicolumn{2}{c}{{One} \texttt{SA}} & \multicolumn{2}{c|}{{Two} \texttt{SA}} & \multicolumn{2}{c}{{Zero} \texttt{SA}} & \multicolumn{2}{c}{{One} \texttt{SA}} & \multicolumn{2}{c}{{Two} \texttt{SA}}\\ \cmidrule(lr){2-3} \cmidrule(lr){4-5} \cmidrule(lr){6-7} \cmidrule(lr){8-9} \cmidrule(lr){10-11} \cmidrule(lr){12-13}
{Metric} & {MSE} & {MAE} & {MSE} & {MAE} & {MSE} & {MAE} & {MSE} & {MAE} & {MSE} & {MAE} & {MSE} & {MAE} \\ \midrule
{96}  & \textbf{0.126} & \textbf{0.218} & 0.128 & 0.220 & 0.133 & 0.225 & \textbf{0.283} & 0.340 & 0.284 &\textbf{0.338} &0.284&0.340\\
{192} & \textbf{0.144} & \textbf{0.235} & 0.150 & 0.238 & 0.153 & 0.245 & \textbf{0.319} & \textbf{0.363} & 0.331 & 0.373 &0.324&0.368\\
{336} & \textbf{0.159} & \textbf{0.252} & 0.167 & 0.257 & 0.169 & 0.263 & \textbf{0.351} & \textbf{0.385} & 0.376 & 0.401 & 0.369& 0.400\\
{720} & \textbf{0.194} & \textbf{0.283} & 0.205 & 0.293 & 0.210 & 0.300 & \textbf{0.400} & \textbf{0.414} & 0.429 & 0.437 & 0.442 & 0.445\\
\bottomrule
\end{tabular} }
\vspace{-0.3cm}
\end{table}

%Visualization
\subsection{Explaining Periodic Patterns through Cross-Attention}
\label{sec:toy}

As noted in \Secref{sec:proposed:structure}, in our proposed model, all operations including embedding and multi-head attention are performed independently for each learnable query. In other words, the forecast for a specific horizon does not depend on other horizons. This approach helps us better understand how each prediction is derived. Therefore, in this subsection, we visualize how the proposed model understands the periodic properties.

To provide an easy understanding, we here consider a simple time series forecasting task with data that consists of two independent signals as follows:
\begin{align}
\label{eq:repeat_signal}
\rvx(t) &= \{x_{(t \bmod \tau)}\}^{\infty}_{t=1}, \quad x_i \sim \mathcal{N}(0, 1) \quad (i = 0, 1, \ldots, \tau-1),\\
\label{eq:shock_signal}
\vy(t) &=
\begin{cases}
+k & \text{if } t \equiv 0 \quad \pmod{S} \\
-k & \text{if } t \equiv \frac{1}{2}S \text{ } \pmod{S}.
\end{cases}
\end{align}

For prediction, we use an input sequence length $L=48$ and a forecasting horizon $T=72$ with signals $\rvx(t)$ and $\vy(t)$ are defined with $\tau=24$, $S=8$, and $k=5$. The patch length is set to 4 without overlapping to elucidate the distinct periodic components with 2 attention heads.

\begin{figure}[ht!] 
    \centering
    \subfloat[{Cross-attention score map of head 1} \label{fig:toy_score_map_a}]{%
       \includegraphics[width=0.4\linewidth]{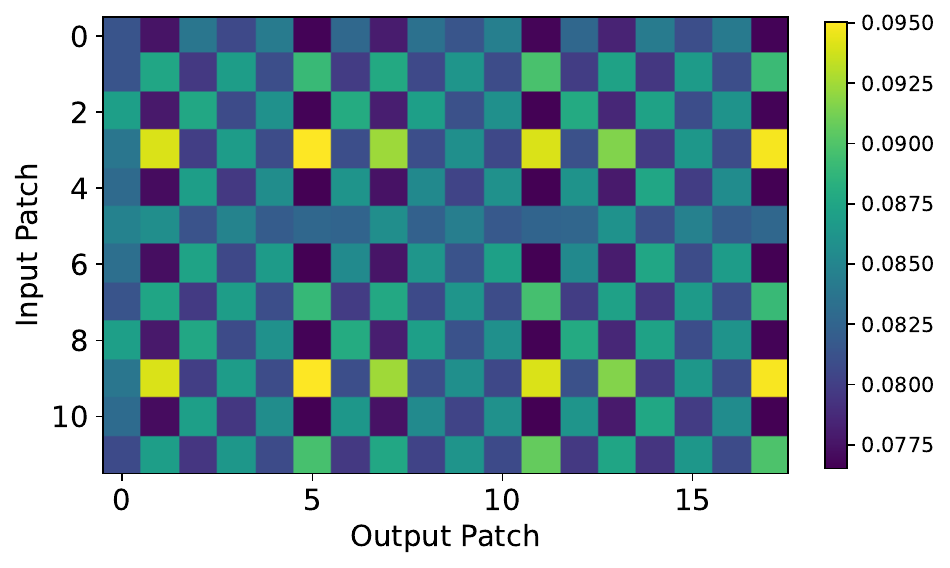}}
    % \quad
    \subfloat[{Cross-attention score map of head 2} \label{fig:toy_score_map_b}]{%
       \includegraphics[width=0.4\linewidth]{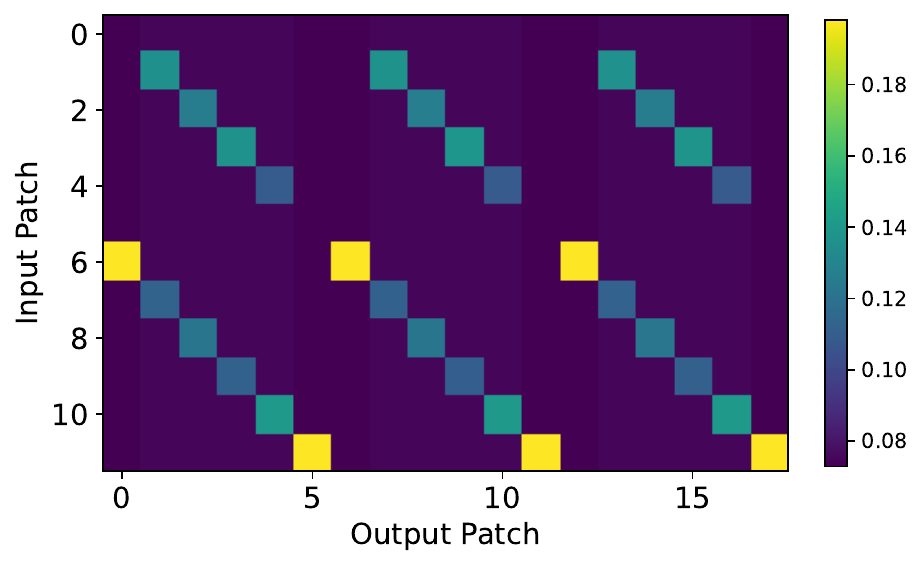}}
    % \quad
    \caption{Score map of cross-attentions between input and output patches.}
    \label{fig:toy_score_map}
\end{figure}

In \Figref{fig:toy_score_map}, we illustrate a score map (12$\times$18) of the cross-attention from the trained CATS. Since both patch length and stride are set to 4, each patch will contain exactly one shock value. We observe that the cross-attentions capture the shocks within the signal and the periodicity of the signal in \Figref{fig:toy_score_map_a} and \Figref{fig:toy_score_map_b}, respectively. \Figref{fig:toy_score_map_a} shows that patches an even number of steps before the current patch contain the shocks of the same direction, resulting in higher attention scores, while odd-numbered steps have lower scores. Moreover, the correlation over 24 steps is clearly demonstrated in patches spaced by multiples of 6 steps, as shown in \Figref{fig:toy_score_map_b}. 
This periodic pattern ensures that the attention mechanism effectively captures the periodicity in $\rvx(t)$, reflecting the model's ability to leverage this periodic information for more accurate predictions. In Appendix, we provide a detailed explanation.

\begin{figure}[ht!] 
    \centering
    \subfloat[{Forecasting results} \label{fig:real_results}]{%
       \includegraphics[width=0.4\linewidth]{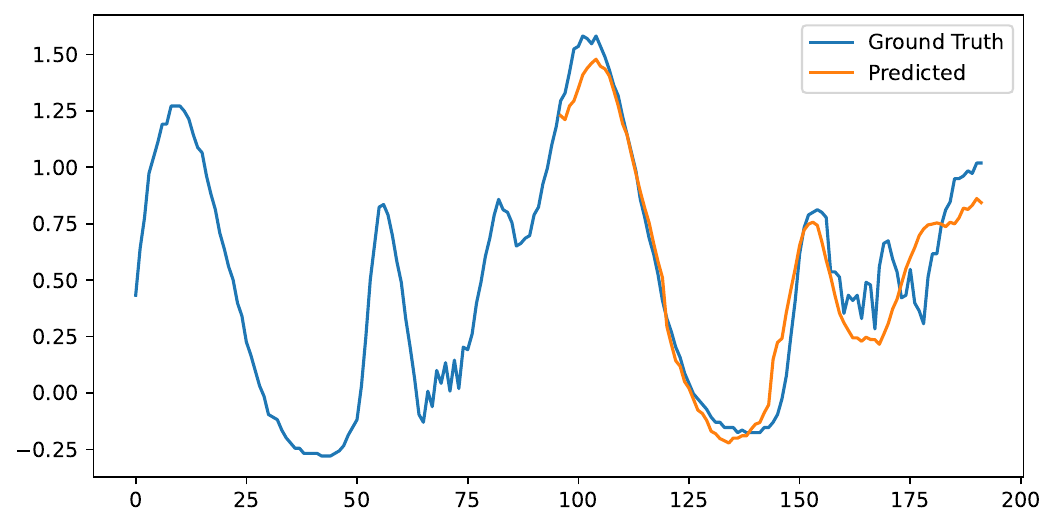}}
    % \quad
    \subfloat[{Averaged score} \label{fig:real_scores}]{%
       \includegraphics[width=0.215\linewidth]{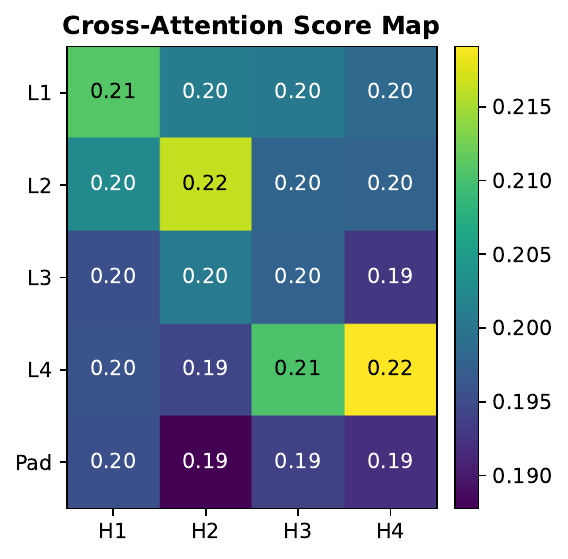}}
    \subfloat[{L2-H2} \label{fig:real_e1}]{%
       \includegraphics[width=0.19\linewidth]{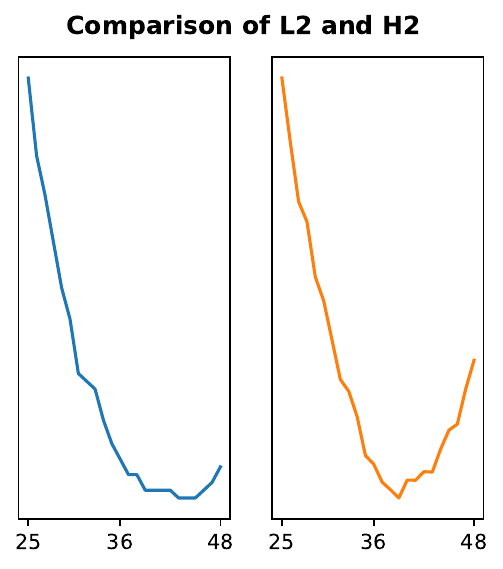}}
    \subfloat[{L4-H4} \label{fig:real_e2}]{%
       \includegraphics[width=0.19\linewidth]{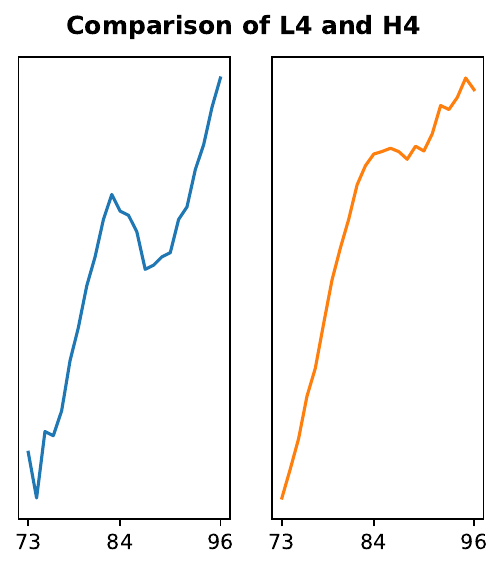}}
    \caption{Illustration of (a) forecasting result, (b) averaged cross-attention score, and (c,d) patches with the highest score on ETTm1. The score map is averaged from all the heads across layers.}

    \label{fig:real_score_map}
\end{figure}

\Figref{fig:real_score_map} illustrates (a) forecasting results, (b) a cross-attention score map (5$\times$4) on the ETTm1 dataset, and (c, d) the two pairs with the highest attention scores. We predict 96 steps with input sequence length 96 on ETTm1. The input patches consist of four patches of 24 lengths and one padding patch. As shown in Fig. \ref{fig:real_e1} and \ref{fig:real_e2}, the patches with high attention scores exhibit similar temporal patterns, demonstrating the ability of CATS to detect sequential and periodic patterns.

\section{Conclusion}

Based on our study, we exploit the advantages of Transformer models in time series forecasting by removing self-attentions and developing a new cross-attention-based architecture. We believe that our model establishes a strong baseline for such forecasting tasks and offers further insights into the complexities of long-term forecasting problems.
Our findings provide a reevaluation of self-attentions in this domain, and we hope that future research can critically assess the efficacy and efficiency across various time series analysis tasks. As a limitation, our proposed methods assume channel independence between variables based on the recent work \cite{nie2023patchtst}. As the time series data in the real-world are highly correlated, we hope future research can address cross-variate dependency with reduced computation complexity based on the proposed architecture.

\section{Acknowledgements}
This work was partly supported by the Institute of Information \& communications Technology Planning \& Evaluation (IITP) grant funded by the Korea government (MSIT) (No. RS-2022-II220984, Development of Artificial Intelligence Technology for Personalized Plug-and-Play Explanation and Verification of Explanation) and the National Research Foundation of Korea (NRF) grant funded by the Korean government (MSIT) (No. RS-2024-00338859). This work was also supported by the MSIT(Ministry of Science and ICT), Korea, under the ITRC(Information Technology Research Center) support program (IITP-2024-RS-2024-00438056) supervised by the IITP.

\bibliographystyle{plainnat}
\bibliography{ref}

\newpage
\appendix

\section{Experimental settings}
\label{app:exp_setting}

\subsection{Datasets}

We evaluated the performance using seven datasets commonly used in long-term time series forecasting, including Weather, Traffic, Electricity, ETT (ETTh1, ETTh2, ETTm1, and ETTm2), and M4. These datasets capture a range of periodic characteristics and scenarios that are difficult to predict in the real world, making them highly suitable for tasks, such as long-term time series forecasting, generation, and imputation. Details of these datasets are described in \Tabref{table:dataset_summary}. The M4 dataset are provided \citet{wu2022timesnet}, while the remaining datasets are provided in \citet{wu2021autoformer}.

\begin{table}[ht!]
\centering
\caption{Details of 13 real-world datasets.} \label{table:dataset_summary}
\begin{tabular}{c|ccccc}
\toprule
 & \textbf{Dimension} & \textbf{Frequency} & \textbf{Timesteps} & \textbf{Information} & \textbf{Forecasting Horizon}\\ \midrule
Weather & 21 & 10-min & 52,696 & Weather & (96, 192, 336, 720) \\ 
Electricity & 321 & Hourly & 17,544 & Electricity & (96, 192, 336, 720) \\ 
Traffic & 862 & Hourly & 26,304 & Transportation & (96, 192, 336, 720) \\ 
ETTh1 & 7 & Hourly & 17,420 & Temperature & (96, 192, 336, 720) \\ 
ETTh2 & 7 & Hourly & 17,420 & Temperature & (96, 192, 336, 720) \\ 
ETTm1 & 7 & 15-min & 69,680 & Temperature & (96, 192, 336, 720) \\ 
ETTm2 & 7 & 15-min & 69,680 & Temperature & (96, 192, 336, 720) \\ 
M4-Quartely & 1 & Quartely & 48000 & Finance     & 8 \\ 
M4-Monthly  & 1 & Monthly  & 96000 & Industry    & 18 \\
M4-Yearly   & 1 & Yearly   & 46000 & Demographic & 6   \\ 
M4-Weekly   & 1 & Weekly   & 718   & Macro       & 13  \\
M4-Daily    & 1 & Daily    & 8454  & Micro       & 14 \\ 
M4-Hourly   & 1 & Hourly   & 828   & Other       & 48 \\ 
\bottomrule
\end{tabular}
\end{table}

\subsection{Hyperparameter Settings}

In every experiment in our paper, following \cite{nie2023patchtst}, we fixed the random seed of 2021 to enhance the reproducibility of our experiments. Additionally, following numerous studies in the field of time series forecasting \cite{nie2023patchtst}, we fixed the input sequence length $L=96$. For the forecasting horizon $T$, we also used the widely accepted values, i.e., $[96, 192, 336, 720]$.
For our model, in all configurations, we adopt the GeGLU activation function \citep{shazeer2020glu} between the two linear layers in the feed-forward network for our model. Additionally, we use learnable positional embedding parameters for the input data and omit positional embeddings for learnable queries to avoid redundant parameter learning. 

For the experiments summarized in \Tabref{table:result_96} and \Tabref{table:full_result_96}, our model uses three cross-attention layers with embedding size $D = 256$, number of attention heads $H=32$. Specifically, to avoid overfitting on small datasets \cite{nie2023patchtst}, we use patch length 48 on the ETTh1 and ETTh2 datasets. Further details on the hyperparameter settings for these experiments are provided in \Tabref{tab:96_params}.

\begin{table}[ht!]
\setlength{\tabcolsep}{3pt}
\centering
\caption{Experimental settings used in \Tabref{table:result_96} and \Tabref{table:full_result_96}.}
\label{tab:96_params}
\resizebox{0.9\columnwidth}{!}{%
\begin{tabular}{c|c|c|c|c|c|c|c}
\toprule [1.2pt]
Metric         & Layers & Embedding Size & Query Sharing & Input Sequence Length & Batch Size &  Epoch & Learning Rate \\ \midrule
                
Weather      & 3 & 256 & False & 96 & 64  & 30  & $10^{-3}$ \\ \midrule
Electricity  & 3 & 256 & False & 96 & 32  & 30  & $10^{-3}$ \\ \midrule
Traffic      & 3 & 256 & True & 96 & 32  & 100 & $10^{-3}$  \\ \midrule
ETTh1        & 3 & 256 & False & 96 & 256 & 10  & $10^{-3}$ \\ \midrule
ETTh2        & 3 & 256 & True & 96 & 256 & 10  & $10^{-3}$  \\ \midrule
ETTm1        & 3 & 256 & False & 96 & 128 & 30  & $10^{-3}$ \\ \midrule
ETTm2        & 3 & 256 & True & 96 & 128 & 30  & $10^{-3}$  \\ \bottomrule[1.2pt]
\end{tabular} %
}
\end{table}

Additionally, for the short-term forecasting experiments on the M4 dataset, we employed a slightly different configuration to better suit the nature of short-term time series. The hyperparameter settings for these experiments are detailed in \Tabref{tab:m4_params}. The complete results for these short-term experiments are presented in \Tabref{table:full_result_m4}, while the full results for the long-term forecasting experiments with a fixed 96 input sequence length are provided in the Appendix in \Tabref{table:full_result_96}.

\begin{table}[ht!]
\vspace{-0.3cm}
\setlength{\tabcolsep}{3pt}
\centering
\caption{Experimental settings of M4 Dataset.}
\label{tab:m4_params}
\resizebox{0.9\columnwidth}{!}{%
\begin{tabular}{l|c|c|c|c|c|c|c|c}
\toprule [1.2pt]
\multicolumn{1}{c|}{Dataset} & Layers & Embedding Size & Input Sequence Length & Batch Size & Epoch & Patience & Learning Rate \\ \midrule
                
M4-Quartely & 3 & 64 & 16 & 32 & 30 &10 & $10^{-3}$ \\ \midrule
M4-Monthly  & 3 & 64 & 36 & 32 & 30 &10 & $10^{-3}$ \\ \midrule
M4-Yearly   & 3 & 64 & 12 & 32 & 30 &10 & $10^{-3}$ \\ \midrule
M4-Weekly   & 3 & 64 & 26 & 32 & 30 &10 & $10^{-3}$ \\ \midrule
M4-Daily    & 3 & 64 & 28 & 32 & 30 &10 & $10^{-3}$ \\ \midrule
M4-Hourly   & 3 & 128& 96 & 32 & 30 &10 & $10^{-3}$ \\ \bottomrule[1.2pt]
\end{tabular} %
}
\vspace{-0.3cm}
\end{table}

\begin{table}[ht!]
\vspace{-0.3cm}
\setlength{\tabcolsep}{3pt}
\centering
\caption{Multivariate long-term forecasting results with recent forecasting models and ours for unified hyperparameter settings. The best results are in \textbf{bold} and the second best are \underline{underlined}.}
\label{table:full_result_96}
\resizebox{1\columnwidth}{!}{%
\begin{tabular}{c|c|cc|cc|cc|cc|cc|cc|cc|cc|cc|cc}
\toprule [1.2pt]
\multicolumn{2}{c}{Models}
&\multicolumn{2}{c}{CATS}&\multicolumn{2}{c}{TimeMixer}&\multicolumn{2}{c}{PatchTST}&\multicolumn{2}{c}{Timesnet}&\multicolumn{2}{c}{Crossformer} & \multicolumn{2}{c}{MICN} & \multicolumn{2}{c}{FiLM}&\multicolumn{2}{c}{DLinear} & \multicolumn{2}{c}{Autoformer}& \multicolumn{2}{c}{Informer}\\ 
\cmidrule(lr){3-4}\cmidrule(lr){5-6}\cmidrule(lr){7-8}\cmidrule(lr){9-10}\cmidrule(lr){11-12}\cmidrule(lr){13-14}\cmidrule(lr){15-16}\cmidrule(lr){17-18}\cmidrule(lr){19-20}\cmidrule(lr){21-22}

\multicolumn{2}{c}{Metric}  & MSE & MAE & MSE & MAE & MSE & MAE & MSE & MAE & MSE & MAE & MSE & MAE & MSE & MAE & MSE & MAE & MSE & MAE & MSE & MAE \\ \midrule

\multirow{4}{*}{\begin{turn}{90}Weather\end{turn}}

 & 96  & \textbf{0.161} & \textbf{0.207} & \underline{0.163} & \underline{0.209} & 0.186 & 0.227 & 0.172 & 0.220 & 0.195 & 0.271 & 0.198 & 0.261 & 0.195 & 0.236 & 0.195 & 0.252  & 0.266 & 0.336 & 0.300 & 0.384 \\
 & 192 & \textbf{0.208} & \textbf{0.250} & \textbf{0.208} & \textbf{0.250} & 0.234 & 0.265 & 0.219 & \underline{0.261} & \underline{0.209} & 0.277 & 0.239 & 0.299 & 0.239 & 0.271 & 0.237 & 0.295  & 0.307 & 0.367 & 0.598 & 0.544 \\
 & 336 & 0.264 & \underline{0.290} & \underline{0.251} & \textbf{0.287} & 0.284 & 0.301 & \textbf{0.246} & 0.337 & 0.273 & 0.332 & 0.285 & 0.336 & 0.289 & 0.306 & 0.282 & 0.331  & 0.359 & 0.395 & 0.578 & 0.523 \\
 & 720 & \underline{0.342} & \textbf{0.341} & \textbf{0.339} & \textbf{0.341} & 0.356 & \underline{0.349} & 0.365 & 0.359 & 0.379 & 0.401 & 0.351 & 0.388 & 0.361 & 0.351 & 0.345 & 0.382  & 0.419 & 0.428 & 1.059 & 0.741 \\ \midrule

\multirow{4}{*}{\begin{turn}{90}Electricity\end{turn}}
 & 96  & \textbf{0.149} & \textbf{0.237} & \underline{0.153} & \underline{0.247} & 0.190 & 0.296 & 0.168 & 0.272 & 0.219 & 0.314 & 0.180 & 0.293 & 0.198 & 0.274 & 0.210 & 0.302  & 0.201 & 0.317 & 0.274 & 0.368 \\
 & 192 & \textbf{0.163} & \textbf{0.250} & \underline{0.166} & \underline{0.256} & 0.199 & 0.304 & 0.184 & 0.322 & 0.231 & 0.322 & 0.189 & 0.302 & 0.198 & 0.278 & 0.210 & 0.305  & 0.222 & 0.334 & 0.296 & 0.386 \\
 & 336 & \textbf{0.180} & \textbf{0.268} & \underline{0.185} & \underline{0.277} & 0.217 & 0.319 & 0.198 & 0.300 & 0.246 & 0.337 & 0.198 & 0.312 & 0.217 & 0.300 & 0.223 & 0.319  & 0.231 & 0.443 & 0.300 & 0.394 \\
 & 720 & \underline{0.219} & \textbf{0.302} & 0.225 & \underline{0.310} & 0.258 & 0.352 & 0.220 & 0.320 & 0.280 & 0.363 & \textbf{0.217} & 0.330 & 0.278 & 0.356 & 0.258 & 0.350  & 0.254 & 0.361 & 0.373 & 0.439 \\ \midrule

\multirow{4}{*}{\begin{turn}{90}Traffic\end{turn}}
 & 96  & \textbf{0.421} & \textbf{0.270} & \underline{0.462} & \underline{0.285} & 0.526 & 0.347 & 0.593 & 0.321 & 0.644 & 0.429 & 0.577 & 0.350 & 0.647 & 0.384 & 0.650 & 0.396  & 0.613 & 0.388 & 0.719 & 0.391 \\
 & 192 & \textbf{0.436} & \textbf{0.275} & \underline{0.473} & \underline{0.296} & 0.522 & 0.332 & 0.617 & 0.336 & 0.665 & 0.431 & 0.589 & 0.356 & 0.600 & 0.361 & 0.598 & 0.370  & 0.616 & 0.382 & 0.696 & 0.379 \\
 & 336 & \textbf{0.453} & \textbf{0.284} & \underline{0.498} & \underline{0.296} & 0.517 & 0.334 & 0.629 & 0.336 & 0.674 & 0.420 & 0.594 & 0.358 & 0.610 & 0.367 & 0.605 & 0.373  & 0.622 & 0.337 & 0.777 & 0.420 \\
 & 720 & \textbf{0.484} & \textbf{0.303} & \underline{0.506} & \underline{0.313} & 0.552 & 0.352 & 0.640 & 0.350 & 0.683 & 0.424 & 0.613 & 0.361 & 0.691 & 0.425 & 0.645 & 0.394  & 0.660 & 0.408 & 0.864 & 0.472 \\ \midrule

\multirow{4}{*}{\begin{turn}{90}ETTm1\end{turn}}
 & 96  & \textbf{0.318} & \textbf{0.357} & \underline{0.320} & \textbf{0.357} & 0.352 & 0.374 & 0.338 & 0.375 & 0.404 & 0.426 & 0.365 & 0.387 & 0.353 & \underline{0.370} & 0.346 & 0.374  & 0.505 & 0.475 & 0.672 & 0.571 \\
 & 192 & \textbf{0.357} & \textbf{0.377} & \underline{0.361} & \underline{0.381} & 0.390 & 0.393 & 0.374 & 0.387 & 0.450 & 0.451 & 0.403 & 0.408 & 0.389 & 0.387 & 0.382 & 0.391  & 0.553 & 0.496 & 0.795 & 0.669 \\
 & 336 & \textbf{0.387} & \textbf{0.401} & \underline{0.390} & \underline{0.404} & 0.421 & 0.414 & 0.410 & 0.411 & 0.532 & 0.515 & 0.436 & 0.431 & 0.421 & 0.408 & 0.415 & 0.415  & 0.621 & 0.537 & 1.212 & 0.871 \\
 & 720 & \textbf{0.448} & \textbf{0.437} & \underline{0.454} & \underline{0.441} & 0.462 & 0.449 & 0.478 & 0.450 & 0.666 & 0.589 & 0.489 & 0.462 & 0.481 & 0.441 & 0.473 & 0.451  & 0.671 & 0.561 & 1.166 & 0.823 \\ \midrule

\multirow{4}{*}{\begin{turn}{90}ETTm2\end{turn}}
 & 96  & \underline{0.178} & \underline{0.261} & \textbf{0.175} & \textbf{0.258} & 0.183 & 0.270 & 0.187 & 0.267 & 0.287 & 0.366 & 0.197 & 0.296 & 0.183 & 0.266 & 0.193 & 0.293 & 0.255 & 0.339 & 0.365 & 0.453 \\
 & 192 & \underline{0.248} & \underline{0.308} & \textbf{0.237} & \textbf{0.299} & 0.255 & 0.314 & 0.249 & 0.309 & 0.414 & 0.492 & 0.284 & 0.361 & 0.248 & 0.305 & 0.284 & 0.361  & 0.281 & 0.340 & 0.533 & 0.563 \\
 & 336 & \underline{0.304} & \underline{0.343} & \textbf{0.298} & \textbf{0.340} & 0.309 & 0.347 & 0.321 & 0.351 & 0.597 & 0.542 & 0.381 & 0.429 & 0.309 & 0.343 & 0.382 & 0.429  & 0.339 & 0.372 & 1.363 & 0.887 \\
 & 720 & \underline{0.402} & \underline{0.402} & \textbf{0.391} & \textbf{0.396} & 0.412 & 0.404 & 0.408 & 0.403 & 1.730 & 1.042 & 0.549 & 0.522 & 0.410 & 0.400 & 0.558 & 0.525 & 0.433 & 0.432 & 3.379 & 1.338 \\ \midrule

\multirow{4}{*}{\begin{turn}{90}ETTh1\end{turn}}
 & 96  & \textbf{0.371} & \textbf{0.395} & \underline{0.375} & \underline{0.400} & 0.460 & 0.447 & 0.384 & 0.402 & 0.423 & 0.448 & 0.426 & 0.446 & 0.438 & 0.433 & 0.397 & 0.412  & 0.449 & 0.459 & 0.865 & 0.713 \\
 & 192 & \textbf{0.426} & \underline{0.422} & \underline{0.429} & \textbf{0.421} & 0.512 & 0.477 & 0.436 & 0.429 & 0.471 & 0.474 & 0.454 & 0.464 & 0.493 & 0.466 & 0.446 & 0.441  & 0.500 & 0.482 & 1.008 & 0.792 \\
 & 336 & \textbf{0.437} & \textbf{0.432} & \underline{0.484} & \underline{0.458} & 0.546 & 0.496 & 0.638 & 0.469 & 0.570 & 0.546 & 0.493 & 0.487 & 0.547 & 0.495 & 0.489 & 0.467  & 0.521 & 0.496 & 1.107 & 0.809 \\
 & 720 & \textbf{0.474} & \textbf{0.461} & \underline{0.498} & \underline{0.482} & 0.544 & 0.517 & 0.521 & 0.500 & 0.653 & 0.621 & 0.526 & 0.526 & 0.586 & 0.538 & 0.513 & 0.510  & 0.514 & 0.512 & 1.181 & 0.865 \\ \midrule

\multirow{4}{*}{\begin{turn}{90}ETTh2\end{turn}}
 & 96  & \textbf{0.287} & \textbf{0.341} & \underline{0.289} & \textbf{0.341} & 0.308 & \underline{0.355} & 0.340 & 0.374 & 0.745 & 0.584 & 0.372 & 0.424 & 0.322 & 0.364 & 0.340 & 0.394  & 0.346 & 0.388 & 3.755 & 1.525 \\
 & 192 & \textbf{0.361} & \textbf{0.388} & \underline{0.372} & \underline{0.392} & 0.393 & 0.405 & 0.402 & 0.414 & 0.877 & 0.656 & 0.492 & 0.492 & 0.404 & 0.414 & 0.482 & 0.479  & 0.456 & 0.453 & 5.602 & 1.931 \\
 & 336 & \textbf{0.374} & \textbf{0.403} & \underline{0.386} & \underline{0.414} & 0.427 & 0.436 & 0.452 & 0.452 & 1.043 & 0.731 & 0.607 & 0.555 & 0.435 & 0.445 & 0.591 & 0.541  & 0.482 & 0.486 & 4.721 & 1.835 \\
 & 720 & \textbf{0.412} & \textbf{0.433} & \textbf{0.412} & \underline{0.434} & \underline{0.436} & 0.450  & 0.462 & 0.468 & 1.104 & 0.763 & 0.824 & 0.655 & 0.447 & 0.458 & 0.839 & 0.661  & 0.515 & 0.511 & 3.647 & 1.625 \\  \bottomrule[1.2pt]

\end{tabular} 
}
\vspace{-0.3cm}
\end{table}

\begin{table}[ht!]
\setlength{\tabcolsep}{3pt}
\centering
\caption{Full results of univariate short-term forecasting in the M4 dataset. All forecasting horizons are in [6, 48]. The best results are in \textbf{bold} and the second best are \underline{underlined}.}
\label{table:full_result_m4}
\resizebox{0.8\columnwidth}{!}{%
\begin{tabular}{c|c|ccccccccc}
\toprule [1.2pt]
\multicolumn{2}{c}{Models}
&CATS & TimeMixer & Timesnet & PatchTST & MICN & FiLM & DLinear & Autoformer & Informer\\ \midrule

\multirow{3}{*}{\begin{turn}{90}\fontsize{8}{7}\selectfont{Quarterly}\end{turn}}

 & SMAPE & \textbf{9.979}  & \underline{9.996}  & 10.100 & 10.644 & 15.214 & 12.925 & 12.145 & 11.338 & 11.360 \\
 & MASE  & \textbf{1.164}  & \underline{1.166}  & 1.182  & 1.278  & 1.963  & 1.664  & 1.520  & 1.365  & 1.401  \\
 & OWA   & \underline{0.878}  & \textbf{0.825}  & 0.890  & 0.949  & 1.407  & 1.193  & 1.106  & 1.012  & 1.027  \\ \midrule

\multirow{3}{*}{\begin{turn}{90}\fontsize{9}{7}\selectfont{Monthly}\end{turn}}

 & SMAPE & \textbf{12.557} & \underline{12.605} & 12.670 & 13.399 & 16.943 & 15.407 & 13.514 & 13.958 & 14.062 \\
 & MASE  & \textbf{0.916}  & \underline{0.919}  & 0.933  & 1.031  & 1.442  & 1.298  & 1.037  & 1.103  & 1.141  \\
 & OWA   & \textbf{0.866}  & \underline{0.869}  & 0.878  & 0.949  & 1.265  & 1.144  & 0.956  & 1.002  & 1.024  \\ \midrule

\multirow{3}{*}{\begin{turn}{90}Yearly\end{turn}}

&SMAPE & \underline{13.263} & \textbf{13.206} & 13.387 & 16.463 & 25.022 & 17.431 & 16.965 & 13.974 & 14.727 \\
&MASE  & \underline{2.967} & \textbf{2.916}  & 2.996  & 3.967  & 7.162  & 4.043  & 4.283  & 3.134  & 3.418  \\
&OWA   & \underline{0.779} & \textbf{0.776}  & 0.786  & 1.003  & 1.667  & 1.042  & 1.058  & 0.822  & 0.881  \\ \midrule

\multirow{3}{*}{\begin{turn}{90}Others\end{turn}}

 & SMAPE & \textbf{4.560}   & \underline{4.564}  & 4.891  & 6.558  & 41.985 & 7.134  & 6.709  & 5.485  & 24.460 \\
 & MASE  & \textbf{3.107}  & \underline{3.115}  & 3.302  & 4.511  & 62.734 & 5.09   & 4.953  & 3.865  & 20.960 \\
 & OWA   & \textbf{0.970}   & \underline{0.982}  & 1.035  & 1.401  & 14.313 & 1.553  & 1.487  & 1.187  & 5.879  \\ \midrule

\multirow{3}{*}{\begin{turn}{90}Average\end{turn}}

 & SMAPE & \textbf{11.701} & \underline{11.723} & 11.829 & 13.152 & 19.638 & 14.863 & 13.639 & 12.909 & 14.086 \\
 & MASE  & \textbf{1.557}  & \underline{1.559}  & 1.585  & 1.945  & 5.947  & 2.207  & 2.095  & 1.771  & 2.718  \\
 & OWA   & \textbf{0.838}  & \underline{0.840}  & 0.851  & 0.998  & 2.279  & 1.125  & 1.051  & 0.939  & 1.230  \\ \bottomrule[1.2pt]
\end{tabular} 
}
\end{table}

\section{Additional Experimental Results}

\subsection{Performance with Longer Input Sequences} \label{app:512_input}

In \Tabref{table:ettm1_param}, we compared models with the number of parameters, GPU memory consumption, and MSE across different input lengths on ETTm1 with varying input sequence lengths. In this section, we provide comprehensive results on longer input sequence lengths $L=512$. The detailed parameters can be found in \Tabref{tab:512_params} and the corresponding experimental results are summarized in \Tabref{tab:main_512}.  
As with unified hyperparameter settings, we follow the settings of the most recent work \citep{wang2024timemixer} to ease comparison.
Overall, the experimental results clearly illustrate the superiority of CATS over recent forecasting models across multiple datasets and prediction horizons. CATS consistently shows the lowest Mean Squared Error (MSE) and Mean Absolute Error (MAE) across a variety of datasets and forecast horizons. For instance, on Electricity, at the 96 forecast horizon, CATS achieves the best MSE score of 0.144 and the best MAE score of 0.189, underscoring its accuracy in predicting electrical demand.

\begin{table}[ht]
\setlength{\tabcolsep}{3pt}
\centering
\caption{Experimental settings with an input sequence length of 512.}
\label{tab:512_params}
\resizebox{0.9\columnwidth}{!}{%
\begin{tabular}{c|c|c|c|c|c|c|c}
\toprule [1.2pt]

Metric         & Layers & Embedding Size & Query Sharing & Input Sequence Length & Batch Size &  Epoch & Learning Rate \\ \midrule
                
Weather      & 3 & 128 & False & 512 & 128  & 30  & $10^{-3}$ \\ \midrule
Electricity  & 3 & 128 & False & 512 & 32  & 30  & $10^{-3}$ \\ \midrule
Traffic      & 3 & 128 & True & 512 & 32  & 100 & $10^{-3}$  \\ \midrule
ETTh1        & 3 & 256 & False & 512 & 128 & 10  & $10^{-3}$ \\ \midrule
ETTh2        & 3 & 256 & True & 512 & 256 & 10  & $10^{-3}$  \\ \midrule
ETTm1        & 3 & 128 & False & 512 & 128 & 30  & $10^{-3}$ \\ \midrule
ETTm2        & 3 & 256 & True & 512 & 128 & 30  & $10^{-3}$  \\ \bottomrule[1.2pt]
\end{tabular}%
}
\end{table} 

\begin{table}[ht!]
\setlength{\tabcolsep}{3pt}
\centering
\caption{Multivariate long-term forecasting results with an input sequence length of 512. The best results are in \textbf{bold} and the second best are \underline{underlined}.}
\label{tab:main_512}
\resizebox{1\columnwidth}{!}{%
\begin{tabular}{c|c|cc|cc|cc|cc|cc|cc|cc|cc|cc|cc}
\toprule [1.2pt]
\multicolumn{2}{c}{Models}
&\multicolumn{2}{c}{CATS}&\multicolumn{2}{c}{TimeMixer}&\multicolumn{2}{c}{PatchTST}&\multicolumn{2}{c}{Timesnet}&\multicolumn{2}{c}{Crossformer} & \multicolumn{2}{c}{MICN} & \multicolumn{2}{c}{FiLM}&\multicolumn{2}{c}{DLinear} & \multicolumn{2}{c}{Autoformer}& \multicolumn{2}{c}{Informer}\\ 
\cmidrule(lr){3-4}\cmidrule(lr){5-6}\cmidrule(lr){7-8}\cmidrule(lr){9-10}\cmidrule(lr){11-12}\cmidrule(lr){13-14}\cmidrule(lr){15-16}\cmidrule(lr){17-18}\cmidrule(lr){19-20}\cmidrule(lr){21-22}

\multicolumn{2}{c}{Metric}  & MSE & MAE & MSE & MAE & MSE & MAE & MSE & MAE & MSE & MAE & MSE & MAE & MSE & MAE & MSE & MAE & MSE & MAE & MSE & MAE\\ \midrule

\multirow{4}{*}{\begin{turn}{90}Weather\end{turn}}

& 96&\textbf{0.144}&0.199& 0.147 & \textbf{0.197} &  0.149 & \underline{0.198} & 0.172 & 0.220 & 0.232 & 0.302 & 0.161 & 0.229 & 0.199 & 0.262 & 0.176 & 0.237  & 0.266 & 0.336 & 0.300 & 0.384 \\
& 192&\textbf{0.188}&\underline{0.240}&\underline{0.189} & \textbf{0.239} &  0.194 & 0.241 & 0.219 & 0.261 & 0.371 & 0.410 & 0.220 & 0.281 & 0.228 & 0.288 & 0.220 & 0.282  & 0.307 & 0.367 & 0.598 & 0.544 \\
& 336&\textbf{0.238}&\textbf{0.280}&\underline{0.241} & \textbf{0.280} &  0.306 & \underline{0.282} & 0.246 & 0.337 & 0.495 & 0.515 & 0.278 & 0.331 & 0.267 & 0.323 & 0.265 & 0.319  & 0.359 & 0.395 & 0.578 & 0.523 \\
& 720 &\textbf{0.308} & \textbf{0.329} &\underline{0.310} & \underline{0.330} &  0.314 & 0.334 & 0.365 & 0.359 & 0.526 & 0.542 & 0.311 & 0.356 & 0.319 & 0.361 & 0.323 & 0.362 & 0.419 & 0.428 & 0.590 & 0.741\\ \midrule

\multirow{4}{*}{\begin{turn}{90}Electricity\end{turn}}
&96  & \textbf{0.126} & \textbf{0.218} & \underline{0.129} & 0.224 & \underline{0.129} & \underline{0.222} & 0.168 & 0.272 & 0.150 & 0.251 & 0.164 & 0.269 & 0.154 & 0.267 & 0.140 & 0.237  & 0.201 & 0.317 & 0.274 & 0.368 \\
&192 & \underline{0.144} & \underline{0.235} & \textbf{0.140} & \textbf{0.220} & 0.147 & 0.240 & 0.184 & 0.322 & 0.161 & 0.260 & 0.177 & 0.285 & 0.164 & 0.258 & 0.153 & 0.249  & 0.222 & 0.334 & 0.296 & 0.386 \\
&336 & \textbf{0.159} & \textbf{0.252} & \underline{0.161} & \underline{0.255} & 0.163 & 0.259 & 0.198 & 0.300 & 0.182 & 0.281 & 0.193 & 0.304 & 0.188 & 0.283 & 0.169 & 0.267  & 0.231 & 0.338 & 0.300 & 0.394 \\
&720 & \textbf{0.194} & \textbf{0.283} & \textbf{0.194} & \underline{0.287} & \underline{0.197} & 0.290 & 0.220 & 0.320 & 0.251 & 0.339 & 0.212 & 0.321 & 0.236 & 0.332 & 0.203 & 0.301 & 0.254 & 0.361 & 0.373 & 0.439 \\ \midrule

\multirow{4}{*}{\begin{turn}{90}Traffic\end{turn}}
 & 96  & \textbf{0.352} & \textbf{0.243} & \underline{0.360} & \underline{0.249} & \underline{0.360} & \underline{0.249} & 0.593 & 0.321 & 0.514 & 0.267 & 0.519 & 0.309 & 0.416 & 0.294 & 0.410 & 0.282  & 0.613 & 0.388 & 0.719 & 0.391 \\
 & 192 & \textbf{0.373} & \underline{0.253} & \underline{0.375} & \textbf{0.250} & 0.379 & 0.256 & 0.617 & 0.336 & 0.549 & 0.252 & 0.537 & 0.315 & 0.408 & 0.288 & 0.423 & 0.287  & 0.616 & 0.382 & 0.696 & 0.379 \\
 & 336 & \underline{0.387} & \textbf{0.260} & \textbf{0.385} & 0.270 & 0.392 & \underline{0.264} & 0.629 & 0.336 & 0.530 & 0.300 & 0.534 & 0.313 & 0.425 & 0.298 & 0.436 & 0.296  & 0.622 & 0.337 & 0.777 & 0.420 \\
 & 720 & \textbf{0.425} & \textbf{0.281} & \underline{0.430} & \textbf{0.281} & 0.432 & \underline{0.286} & 0.640 & 0.350 & 0.573 & 0.313 & 0.577 & 0.325 & 0.520 & 0.353 & 0.466 & 0.315 & 0.660 & 0.408 & 0.864 & 0.472 \\ \midrule

\multirow{4}{*}{\begin{turn}{90}ETTh1\end{turn}}
 & 96  & 0.373 & 0.401 & \textbf{0.361} & \textbf{0.390} & \underline{0.370} & \underline{0.400} & 0.384 & 0.402 & 0.418 & 0.438 & 0.421 & 0.431 & 0.422 & 0.432 & 0.375 & 0.399  & 0.449 & 0.459 & 0.865 & 0.713 \\
 & 192 & \textbf{0.401} & 0.421 & 0.409 & \textbf{0.414} & 0.413 & 0.429 & 0.436 & 0.429 & 0.539 & 0.517 & 0.474 & 0.487 & 0.462 & 0.458 & \underline{0.405} & \underline{0.416}  & 0.500 & 0.482 & 0.080 & 0.792 \\
 & 336 & \textbf{0.415} & \underline{0.434} & 0.430 & \textbf{0.429} & \underline{0.422} & 0.440 & 0.638 & 0.469 & 0.709 & 0.638 & 0.569 & 0.551 & 0.501 & 0.483 & 0.439 & 0.443 & 0.521 & 0.496 & 0.107 & 0.809 \\
 & 720 & \textbf{0.435} & \textbf{0.446} & \underline{0.445} & \underline{0.460} & 0.447 & 0.468 & 0.521 & 0.500 & 0.733 & 0.636 & 0.770 & 0.672 & 0.544 & 0.526 & 0.472 & 0.490 & 0.514 & 0.512 & 0.181 & 0.865 \\ \midrule

\multirow{4}{*}{\begin{turn}{90}ETTh2\end{turn}}
 & 96  & \textbf{0.256} & \textbf{0.328} & 0.271 & 0.330 & 0.274 & 0.337 & 0.340 & 0.374 & 0.425 & 0.463 & 0.299 & 0.364 & 0.323 & 0.370 & 0.289 & 0.353  & 0.358 & 0.397 & 0.755 & 0.525 \\
 & 192 & \underline{0.311} & \underline{0.366} & 0.317 & 0.402 & 0.314 & 0.382 & \textbf{0.231} & \textbf{0.322} & 0.473 & 0.500 & 0.441 & 0.454 & 0.391 & 0.415 & 0.383 & 0.418  & 0.456 & 0.453 & 0.602 & 0.931 \\
 & 336 & \textbf{0.319} & \textbf{0.382} & 0.332 & 0.396 & \underline{0.329} & \underline{0.384} & 0.452 & 0.452 & 0.581 & 0.562 & 0.654 & 0.567 & 0.415 & 0.440 & 0.448 & 0.465  & 0.482 & 0.486 & 0.721 & 0.835 \\
 & 720 & 0.395 & 0.438 & \textbf{0.342} & \textbf{0.408} & \underline{0.379} & \underline{0.422} & 0.462 & 0.468 & 0.775 & 0.665 & 0.956 & 0.716 & 0.441 & 0.459 & 0.605 & 0.551 & 0.515 & 0.511 & 0.647 & 0.625 \\ \midrule

\multirow{4}{*}{\begin{turn}{90}ETTm1\end{turn}}
 & 96  & \textbf{0.283} & \textbf{0.340} & \underline{0.291} & \textbf{0.340} & 0.293 & 0.346 & 0.338 & 0.375 & 0.361 & 0.403 & 0.316 & 0.362 & 0.302 & 0.345 & 0.299 & \underline{0.343} & 0.505 & 0.475 & 0.672 & 0.571 \\
 & 192 & \textbf{0.319} & \textbf{0.363} & \underline{0.327} & \underline{0.365} & 0.333 & 0.370 & 0.374 & 0.387 & 0.387 & 0.422 & 0.363 & 0.390 & 0.338 & 0.368 & 0.335 & 0.365 & 0.553 & 0.496 & 0.795 & 0.669 \\
 & 336 & \textbf{0.351} & \underline{0.385} & \underline{0.360} & \textbf{0.381} & 0.369 & 0.392 & 0.410 & 0.411 & 0.605 & 0.572 & 0.408 & 0.426 & 0.373 & 0.388 & 0.369 & 0.386 & 0.621 & 0.537 & 0.212 & 0.871 \\
 & 720 & \textbf{0.400} & \textbf{0.414} & \underline{0.415} & \underline{0.417} & 0.416 & 0.420 & 0.478 & 0.450 & 0.703 & 0.645 & 0.481 & 0.476 & 0.420 & 0.420 & 0.425 & 0.421 & 0.671 & 0.561 & 0.166 & 0.823 \\ \midrule

\multirow{4}{*}{\begin{turn}{90}ETTm2\end{turn}}
 & 96  & \underline{0.165} & \underline{0.256} & \textbf{0.164} & \textbf{0.254} & 0.166 & \underline{0.256} & 0.187 & 0.267 & 0.275 & 0.358 & 0.179 & 0.275 & 0.165 & 0.256 & 0.167 & 0.260 & 0.255 & 0.339 & 0.365 & 0.453 \\
 & 192 & \textbf{0.221} & 0.297 & \underline{0.223} & \textbf{0.295} & \underline{0.223} & \underline{0.296} & 0.249 & 0.309 & 0.345 & 0.400 & 0.307 & 0.376 & 0.222 & 0.296 & 0.224 & 0.303 & 0.281 & 0.340 & 0.533 & 0.563 \\
 & 336 & \textbf{0.274} & 0.334 & \underline{0.279} & \underline{0.330} & \textbf{0.274} & \textbf{0.329} & 0.321 & 0.351 & 0.657 & 0.528 & 0.325 & 0.388 & 0.277 & 0.333 & 0.281 & 0.342 & 0.339 & 0.372 & 0.363 & 0.887 \\
 & 720 & \underline{0.362} & 0.390 & \textbf{0.359} & \textbf{0.383} & \underline{0.362} & \underline{0.385} & 0.408 & 0.403 & 0.208 & 0.753 & 0.502 & 0.490 & 0.371 & 0.389 & 0.397 & 0.421 & 0.422 & 0.419 & 0.379 & 0.388 \\ \bottomrule[1.2pt]
\end{tabular} 
}
\end{table}

\subsection{Additional Results for \Secref{sec:efficient}}

\begin{table}[ht!]
\setlength{\tabcolsep}{3pt}
\centering
\caption{Comparison of models with the number of parameters, GPU memory consumption, and MSE across different input sequence lengths on ETTm1. Full results of \Tabref{table:ettm1_param}.}
\label{table:ettm1_details_input}
\begin{tabular}{l|rrrrrrr}
\toprule [1.2pt]

& \multicolumn{7}{c}{Parameters across different input lengths} \\ \cmidrule(lr){2-8}

\textbf{Models}& \multicolumn{1}{c}{96} & \multicolumn{1}{c}{192} & \multicolumn{1}{c}{336} & \multicolumn{1}{c}{512} & \multicolumn{1}{c}{720} & \multicolumn{1}{c}{1440} &  \multicolumn{1}{c}{2880} \\ \midrule

PatchTST & 1,506,384 & 2,612,304 & 4,271,184&6,298,704 & 8,694,864 & 16,989,264 & 33,578,064 \\
TimeMixer &190,313  & 484,217 & 1,129,193 & 2,250,137 & 4,046,633 & 14,211,593 & 52,912,313 \\
DLinear &139,680 & 277,920 & 485,280 & 738,720 & 1,038,240 & 2,075,040 & 4,148,640  \\
CATS & 360,264 & 360,776 & 361,544 & 362,440 & 363,592 & 367,432 & 375,112 \\ \midrule

& \multicolumn{7}{c}{GPU Memory Consumption across different input lengths} \\ \cmidrule(lr){2-8}
\textbf{Models}& \multicolumn{1}{c}{96} & \multicolumn{1}{c}{192} & \multicolumn{1}{c}{336} & \multicolumn{1}{c}{512} & \multicolumn{1}{c}{720} & \multicolumn{1}{c}{1440} &  \multicolumn{1}{c}{2880} \\ \midrule
PatchTST & 2,234MB & 2,650MB & 3,484MB & 4,914MB & 7,368MB & 2,1968MB & 58,590MB \\
TimeMixer &2,204MB & 2,522MB & 2,914MB & 3,414MB & 3,888MB & 5,876MB & 10,324MB\\
DLinear &1,098MB & 1,102MB & 1,104MB & 1,114MB & 1,114MB & 1,154MB & 1,214MB  \\
CATS & 1,712MB & 1,796MB & 1,884MB & 2,042MB & 2,140MB & 2,700MB & 3,826MB \\ \midrule

& \multicolumn{7}{c}{MSE across different input lengths} \\ \cmidrule(lr){2-8}
\textbf{Models}& \multicolumn{1}{c}{96} & \multicolumn{1}{c}{192} & \multicolumn{1}{c}{336} & \multicolumn{1}{c}{512} & \multicolumn{1}{c}{720} & \multicolumn{1}{c}{1440} &  \multicolumn{1}{c}{2880} \\ \midrule
PatchTST & \multicolumn{1}{c}{0.457} & \multicolumn{1}{c}{0.424} & \multicolumn{1}{c}{0.418} & \multicolumn{1}{c}{0.420} & \multicolumn{1}{c}{0.418} & \multicolumn{1}{c}{0.420} & \multicolumn{1}{c}{0.413} \\
TimeMixer & \multicolumn{1}{c}{0.454} & \multicolumn{1}{c}{0.433} & \multicolumn{1}{c}{0.428} & \multicolumn{1}{c}{0.436} & \multicolumn{1}{c}{0.425} & \multicolumn{1}{c}{0.414} & \multicolumn{1}{c}{0.472} \\
DLinear & \multicolumn{1}{c}{0.473} & \multicolumn{1}{c}{0.438} & \multicolumn{1}{c}{0.426} & \multicolumn{1}{c}{0.427} & \multicolumn{1}{c}{0.422} & \multicolumn{1}{c}{0.401} & \multicolumn{1}{c}{0.408} \\
CATS & \multicolumn{1}{c}{\textbf{0.450}} & \multicolumn{1}{c}{\textbf{0.418}} & \multicolumn{1}{c}{\textbf{0.407}} & \multicolumn{1}{c}{\textbf{0.400}} & \multicolumn{1}{c}{\textbf{0.402}} & \multicolumn{1}{c}{\textbf{0.399}} & \multicolumn{1}{c}{\textbf{0.395}} \\ \bottomrule[1.2pt]
\end{tabular}
\end{table}

\begin{table}[ht!]
\setlength{\tabcolsep}{3pt}
\centering
\caption{Comparison of models with the number of parameters, GPU memory consumption, and MSE across different input sequence lengths on Weather.}
\label{table:weather_details_input}
\begin{tabular}{l|rrrrrrr}
\toprule [1.2pt]

& \multicolumn{7}{c}{Parameters across different input lengths} \\ \cmidrule(lr){2-8}

\textbf{Models}& \multicolumn{1}{c}{96} & \multicolumn{1}{c}{192} & \multicolumn{1}{c}{336} & \multicolumn{1}{c}{512} & \multicolumn{1}{c}{720} & \multicolumn{1}{c}{1440} &  \multicolumn{1}{c}{2880} \\ \midrule

PatchTST & 1,506,384 & 2,612,304 & 4,271,184 & 6,298,704 & 8,694,864 & 16,989,264 & 33,578,064 \\
TimeMixer & 219,249 & 595,305 & 1,465,569 & 3,028,185 & 5,582,529 & 20,343,969 & 77,423,049 \\
DLinear & 139,680 & 277,920 & 485,280 & 738,720 & 1,038,240 & 2,075,040 & 4,148,640 \\
CATS & 370,344 & 370,856 & 371,624 & 372,520 & 373,672 & 377,512 & 385,192 \\ \midrule

& \multicolumn{7}{c}{GPU Memory Consumption across different input lengths} \\ \cmidrule(lr){2-8}
\textbf{Models}& \multicolumn{1}{c}{96} & \multicolumn{1}{c}{192} & \multicolumn{1}{c}{336} & \multicolumn{1}{c}{512} & \multicolumn{1}{c}{720} & \multicolumn{1}{c}{1440} &  \multicolumn{1}{c}{2880} \\ \midrule
PatchTST & 1,680MB & 2,110MB & 2,762MB & 4,596MB & 5,726MB & 16,472MB & 45,278MB \\
TimeMixer & 1,894MB & 2,154MB & 2,728MB & 3,414MB & 4,356MB & 8,358MB & 20,624MB \\
DLinear & 1,106MB & 1,114MB & 1,188MB & 1,188MB & 1,188MB & 1,362MB & 1,632MB \\
CATS & 1,522MB & 1,590MB & 1,665MB & 1,755MB & 1,892MB & 2,282MB & 3,140MB \\ \midrule

& \multicolumn{7}{c}{MSE across different input lengths} \\ \cmidrule(lr){2-8}
\textbf{Models}& \multicolumn{1}{c}{96} & \multicolumn{1}{c}{192} & \multicolumn{1}{c}{336} & \multicolumn{1}{c}{512} & \multicolumn{1}{c}{720} & \multicolumn{1}{c}{1440} &  \multicolumn{1}{c}{2880} \\ \midrule
PatchTST & \multicolumn{1}{c}{0.351} & \multicolumn{1}{c}{0.336} & \multicolumn{1}{c}{0.320} & \multicolumn{1}{c}{0.315} & \multicolumn{1}{c}{0.309} & \multicolumn{1}{c}{0.308} & \multicolumn{1}{c}{0.312} \\
TimeMixer & \multicolumn{1}{c}{\textbf{0.339}} & \multicolumn{1}{c}{0.331} & \multicolumn{1}{c}{0.318} & \multicolumn{1}{c}{0.319} & \multicolumn{1}{c}{0.324} & \multicolumn{1}{c}{0.318} & \multicolumn{1}{c}{0.327} \\
DLinear & \multicolumn{1}{c}{0.346} & \multicolumn{1}{c}{0.334} & \multicolumn{1}{c}{0.325} & \multicolumn{1}{c}{0.320} & \multicolumn{1}{c}{0.316} & \multicolumn{1}{c}{0.311} & \multicolumn{1}{c}{0.309} \\
CATS & \multicolumn{1}{c}{0.342} & \multicolumn{1}{c}{\textbf{0.325}} & \multicolumn{1}{c}{\textbf{0.314}} & \multicolumn{1}{c}{\textbf{0.308}} & \multicolumn{1}{c}{\textbf{0.305}} & \multicolumn{1}{c}{\textbf{0.301}} & \multicolumn{1}{c}{\textbf{0.291}} \\

\bottomrule[1.2pt]
\end{tabular}
\end{table}

We provide additional experimental results to support the findings discussed in \Secref{sec:efficient}. Tables \ref{table:ettm1_details_input} and \ref{table:weather_details_input} summarize detailed comparisons of the number of parameters, GPU memory consumption, and MSE across different input lengths for the ETTm1 and Weather datasets, respectively. The linear models, TimeMixer and DLinear, exhibit smaller parameters for shorter input lengths. Despite CATS having slightly more parameters than TimeMixer for smaller inputs, it outperforms in terms of memory usage and MSE. This suggests that CATS is more efficient and effective in handling shorter inputs.
For PatchTST, the number of parameters does not increase within the actual Transformer backbone as the input length increases. However, due to the need to flatten and project all inputs at the end, the parameters scale linearly with the input length. This highlights a limitation of the Encoder’s architecture. On the other hand, TimeMixer's parameters grow almost quadratically as the input length doubles. Similarly, DLinear’s parameters increase linearly with the input length.
Our proposed model, CATS demonstrates significant efficiency through parameter sharing, where the parameters hardly increase with longer inputs. Notably, from an input length of 336, CATS has fewer parameters than DLinear, showcasing the deep learning model's advantage in detecting inherent patterns in the data.

Regarding GPU memory consumption, we observe that both PatchTST and TimeMixer require significantly more GPU memory as the input length increases. For example, PatchTST's GPU memory usage scales drastically, making it less feasible for long input sequences. TimeMixer also shows an increase in GPU memory consumption, although it is less severe than PatchTST.
In contrast, DLinear maintains a relatively constant GPU memory usage, demonstrating its efficiency in terms of computational resources. However, CATS stands out by offering a balanced approach, with moderate GPU memory usage that scales more favorably compared to PatchTST and TimeMixer. This balance between memory efficiency and performance is crucial for practical applications requiring long-term time series forecasting.

Furthermore, when analyzing the MSE across different input lengths, CATS consistently shows the best performance. It maintains lower MSE compared to other models across all input lengths. This robustness in performance, combined with its efficient parameter and memory usage, highlights the superiority of CATS in long-term time series forecasting tasks.
Overall, these results show the advantages of CATS in terms of parameter efficiency, GPU memory consumption, and forecasting accuracy. These findings support the proposed model's potential for practical and scalable time series forecasting solutions.

\Tabref{tab:detail_traffic_large_input} presents the full results on the Traffic dataset. Here, we use the Traffic dataset with a batch size of 8. All GPU memory consumption was measured in a setting using four multi-GPUs. As shown in \Tabref{tab:detail_traffic_large_input}, CATS with a 2880 input sequence length consistently outperforms models with a 512 input sequence length, including PatchTST and TimeMixer. Specifically, CATS demonstrates fewer parameters, lower GPU memory consumption, and faster running speeds. These results highlight the efficiency of CATS with large input sizes. The Traffic dataset, characterized by high-dimensional data, shows a significant reduction in MSE when using longer input sequences.

\Tabref{tab:detail_electricity_large_input} provides the full results on the Electricity dataset. Similar to the Traffic dataset, CATS shows superior efficiency in training, particularly with an input size of 2880, across all cases. Here, we use the Electricity dataset with a batch size of 32. All GPU memory consumption was measured in a setting using four multi-GPUs. In this experiment, CATS with a 512 input sequence length did not use parameter sharing for queries, while CATS with a 2880 input sequence length did. This demonstrates the effectiveness of query parameter sharing when utilizing large amounts of data for training. The results confirm that query sharing among dimensions leads to greater efficiency and improved performance.

\begin{table}[]
\centering
\caption{Comparison of models with the number of parameters, GPU memory consumption, running speed, and MSE across different forecasting horizon sizes on Traffic. Full results of \Figref{fig:efficiency_analysis}.}
\label{tab:detail_traffic_large_input}
\begin{tabular}{c|l|rrrr}
\toprule
\textbf{Horizon}              & \textbf{Models}        & \multicolumn{1}{c}{\textbf{Paramters}} & \multicolumn{1}{c}{\textbf{Gpu Memory}} & \multicolumn{1}{c}{\textbf{Running Time}} & \multicolumn{1}{c}{\textbf{MSE}} \\ \midrule
\multirow{4}{*}{96}  & PatchTST      & 1,186,272   & 28.54GB   & 0.1390s/iter  & 0.360 \\
                     & TimeMixer     & 2,442,961   & 38.12GB    & 0.2548s/iter  & 0.360 \\
                     & CATS ($L=512$)  & 357,496    & 5.81GB     & 0.0533s/iter& 0.352 \\
                     & {CATS ($L=2880$)} & 370,168    & 9.79GB     & 0.1158s/iter & \textbf{0.339} \\ \midrule
\multirow{4}{*}{192} & PatchTST      & 1,972,800   & 28.34GB    & 0.1412s/iter  & 0.379 \\
                     & TimeMixer     & 2,535,505   & 38.13GB    & 0.2596s/iter  & 0.375 \\
                     & CATS ($L=512$)  & 357,592    & 6.73GB     & 0.0571s/iter  & 0.373 \\
                     & {CATS ($L=2880$)} & 370,264    & 11.10GB    & 0.1209s/iter  & \textbf{0.362} \\\midrule
\multirow{4}{*}{336} & PatchTST      & 3,152,592   & 28.91GB    & 0.1487s/iter  & 0.392 \\
                     & TimeMixer     & 2,674,321   & 38.69GB    & 0.2647s/iter   & 0.385 \\
                     & CATS ($L=512$)  & 357,736    & 7.46GB     & 0.0584s/iter  & 0.387 \\
                     & {CATS ($L=2880$)} & 370,408    & 12.72GB    & 0.1266s/iter  & \textbf{0.379} \\\midrule
\multirow{4}{*}{720} & PatchTST      & 6,298,704   & 29.15GB    & 0.1628s/iter  & 0.432 \\
                     & TimeMixer     & 3,044,497   & 41.17GB   & 0.2777s/iter  & 0.430 \\
                     & CATS ($L=512$)  & 358,120    & 10.10GB    & 0.0734s/iter  & 0.423 \\
                     & {CATS ($L=2880$)} & 370,792    & 18.40GB   & 0.1556s/iter   & \textbf{0.420} \\ \bottomrule
\end{tabular}
\end{table}

\begin{table}[!ht]
\centering
\caption{Comparison of models with the number of parameters, GPU memory consumption, running speed, and MSE across different forecasting horizon sizes on Electricity.}
\label{tab:detail_electricity_large_input}
\begin{tabular}{c|l|rrrr}
\toprule
\textbf{Horizon}              & \textbf{Models}        & \multicolumn{1}{c}{\textbf{Paramters}} & \multicolumn{1}{c}{\textbf{Gpu Memory}} & \multicolumn{1}{c}{\textbf{Running Time}} & \multicolumn{1}{c}{\textbf{MSE}} \\ \midrule
\multirow{4}{*}{96}  & PatchTST      & 1,186,272   & 40.36GB   & 0.2021s/iter  & 0.129 \\
                     & TimeMixer     & 2,429,049   & 33.80GB    & 0.2118s/iter  & 0.129 \\
                     & CATS ($L=512$)  & 388,216    & 6.89GB     & 0.0587s/iter& \textbf{0.126} \\
                     & {CATS ($L=2880$)} & 370,168    & 12.82GB     & 0.1653s/iter & \textbf{0.126} \\ \midrule
\multirow{4}{*}{192} & PatchTST      & 1,972,800   & 40.39GB    & 0.2048s/iter  & 0.147 \\
                     & TimeMixer     & 2,521,593   & 33.81GB    & 0.2212s/iter  & 0.140 \\
                     & CATS ($L=512$)  & 419,032    & 8.07GB     & 0.0636s/iter  & 0.144 \\
                     & {CATS ($L=2880$)} & 370,264    & 14.70GB    & 0.1725s/iter  & \textbf{0.139} \\\midrule
\multirow{4}{*}{336} & PatchTST      & 3,152,592   &40.42GB    & 0.2070s/iter  & 0.163 \\
                     & TimeMixer     & 2,660,409   & 34.24GB    & 0.2314s/iter   & 0.161 \\
                     & CATS ($L=512$)  & 465,256    & 9.15GB     & 0.0690s/iter  & 0.159 \\
                     & {CATS ($L=2880$)} & 370,408    & 17.38GB    & 0.1839s/iter  & \textbf{0.153} \\\midrule
\multirow{4}{*}{720} & PatchTST      & 6,298,704   & 41.40GB    & 0.2313s/iter  & 0.197 \\
                     & TimeMixer     & 3,030,585   & 36.13GB   & 0.2478s/iter  & 0.194 \\
                     & CATS ($L=512$)  & 588,520    & 12.77GB    & 0.0964s/iter  & 0.194 \\
                     & {CATS ($L=2880$)} & 370,792    & 25.86GB   & 0.2262s/iter   & \textbf{0.183} \\ \bottomrule
\end{tabular}
\end{table}

\newpage
\subsection{Ablation Study on Query-adaptive Masking} % Robustness of Horizon-adaptive Masking strategy

\begin{figure}[ht!]
    \centering
    \subfloat[{ETTm1 with $T = 720$} \label{fig:ETTm1_masking_ablation}]{%
       \includegraphics[width=0.48\linewidth]{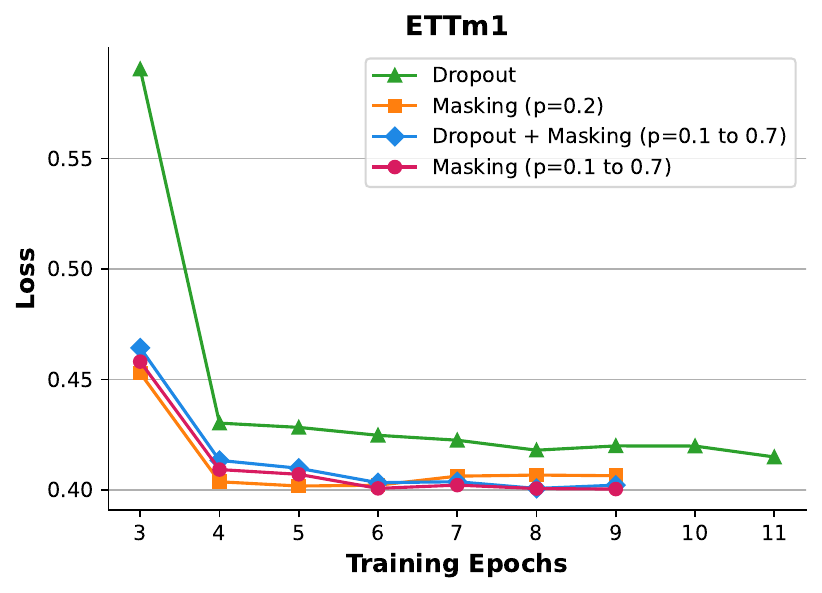}}
    \quad
    \subfloat[{Weather with $T = 720$} \label{fig:Weather_masking_ablation}]{%
       \includegraphics[width=0.48\linewidth]{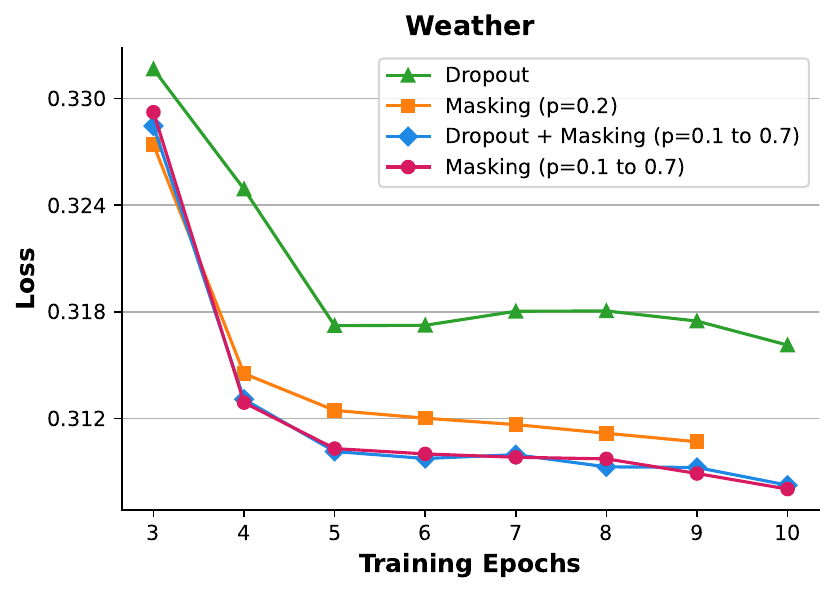}}
    % \quad
    \caption{Comparison of performance with query-adaptive masking with two different probabilities, dropout, and using both query-adaptive masking and dropout. The results of $p=0.1$ to $0.7$ indicate a probability masking that is linearly increased proportionally to the horizon predicted by the query.}
    \label{fig:mask_ablation}
\end{figure}

In this section, we demonstrate the effectiveness of query-adaptive masking compared to dropout, which is a widely adopted technique in transformer-based forecasting models. 
We consider four different setups: using only dropout, using query-adaptive masking with fixed probabilities, query-adaptive masking with linearly increasing probabilities, and using both methods simultaneously. As shown in \Figref{fig:mask_ablation}, the query-adaptive masking shows better forecasting performance and faster converge speed compared to dropout. Applying a gradually increasing masking probability based on the horizon predicted by the query shows slight performance improvements over using a fixed probability or combining with dropout. In contrast, using dropout alone shows noticeable differences in both convergence speed and overall performance. This demonstrates that when multiple inputs with different forecasting horizons share a single model, probabilistic masking is more beneficial for model training than dropout.

\section{Detailed Explanation of \Secref{sec:toy}}
\label{app:toy}
In this section, we provide the detailed results of experiments in \Secref{sec:toy}.
We first restate the formulation of two independent signals used in \Secref{sec:toy} as follows:
\begin{align*}
\rvx(t) &= \{x_{(t \bmod \tau)}\}^{\infty}_{t=1}, \quad x_i \sim \mathcal{N}(0, 1) \quad (i = 0, 1, \ldots, \tau-1),\\
\vy(t) &=
\begin{cases}
+k & \text{if } t \equiv 0 \pmod{S} \\
-k & \text{if } t \equiv \frac{1}{2}S \pmod{S}
\end{cases} ,
\end{align*}

\begin{figure}[ht!]
    \centering
    \subfloat[{Two input signals} \label{fig:toy_signals}]{%
       \includegraphics[width=0.5\linewidth]{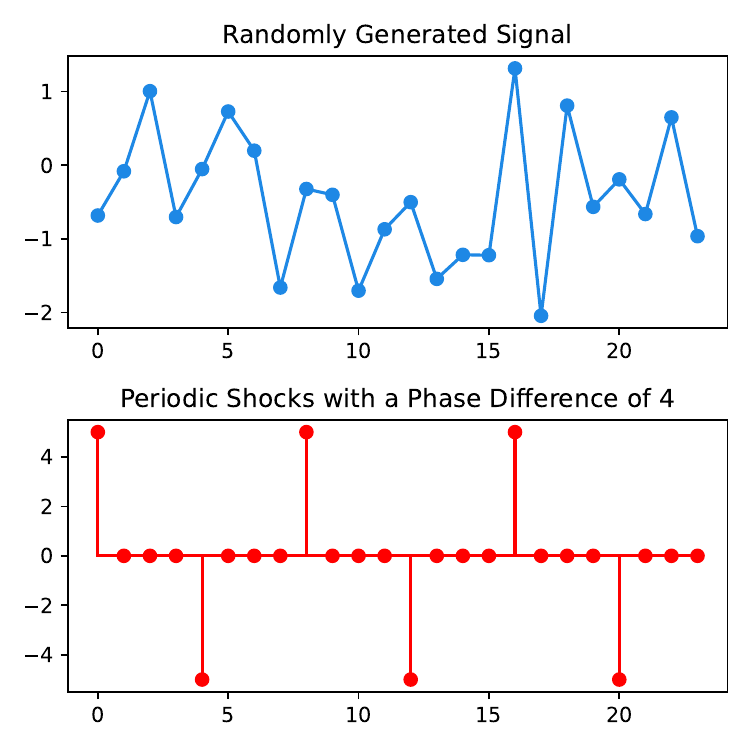}}
    \subfloat[{Sum of two input signals} \label{fig:toy_signal}]{%
       \includegraphics[width=0.5\linewidth]{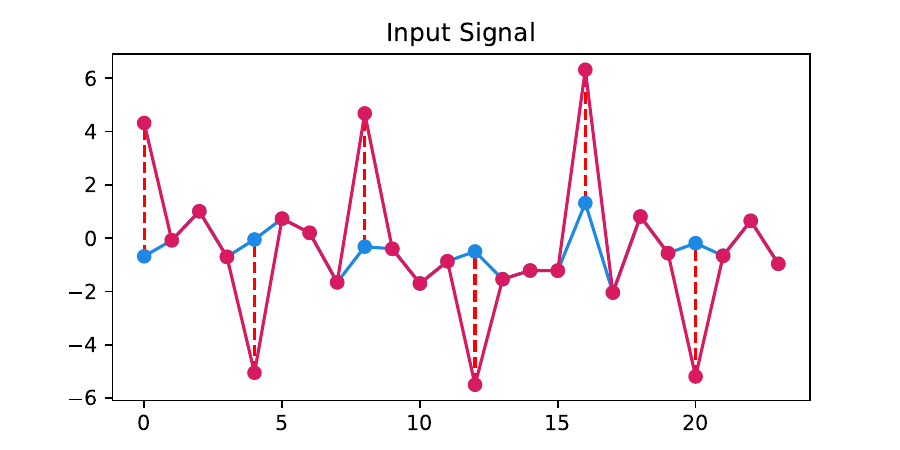}}
    \caption{Visualization of input signals for toy experiment.}
    \label{fig:toy_input_signal}
\end{figure}

We use the model parameters as follows: the patch length is 4 without overlapping, the decoder has 1 layer, and there are 2 attention heads. The signals $\rvx(t)$ and $\vy(t)$ are defined with $\tau=24$, $S=8$, and $k=5$.  The visualization of synthetic data is shown in \Figref{fig:toy_input_signal}. 
We utilize an input sequence length $L=48$ and a forecasting horizon $T=72$. This setup allows us to generate time series data with distinct periodic components.

In the main paper, \Figref{fig:toy_score_map} displays a cross-attention score map between the input patch and the output patch derived from this experiment. The left figure presents the attention score of the first attention head, illustrating the model's ability to detect shocks within the signal. The right figure more clearly demonstrates the periodicity of the signal.
Given that the patch length and stride are both set to 4, each patch will contain exactly one shock value, either -5 or +5. This is because the shocks occur every 4 steps, alternating between positive and negative shocks. Consequently, the patch immediately preceding the current patch will contain a different shock, leading to lower attention scores due to the differing shock values. In contrast, patches that are an even number of steps before the current patch will contain the same type of shock, resulting in higher attention scores. These points are well illustrated in  \Figref{fig:toy_score_map_a}, where the varying attention scores correspond to the presence of alternating shocks. This pattern helps to highlight the alternating shock signal within the data.

Additionally, if there is a correlation with the series preceding 24 steps, the patches that are 6 steps or multiples of 6 steps before the current patch will exhibit high attention scores due to the periodic nature of the signal $\rvx(t)$. The diagonal formation of the attention scores, which accurately follows a period of 24, is clearly depicted in \Figref{fig:toy_score_map_b}, highlighting the model's capability to utilize fixed-period input patches to predict future outcomes. This periodic pattern ensures that the attention mechanism effectively captures the 24-step periodicity in $\rvx(t)$, reflecting the model's ability to leverage this periodic information for more accurate predictions.

This experimental configuration provides a robust framework to evaluate how well our proposed model captures and interprets the underlying patterns in the data, specifically focusing on the alternating shock signal and the periodic nature of the normal signal. This dual emphasis on both the shock signal and the periodicity of the normal signal enhances the interpretability and predictive performance of the model, distinctly demonstrating how the model leverages periodic information to enhance prediction accuracy.

\begin{figure}[ht!] 
    \centering
    \subfloat[{Forecasting results}]{%
       \includegraphics[width=0.4\linewidth]{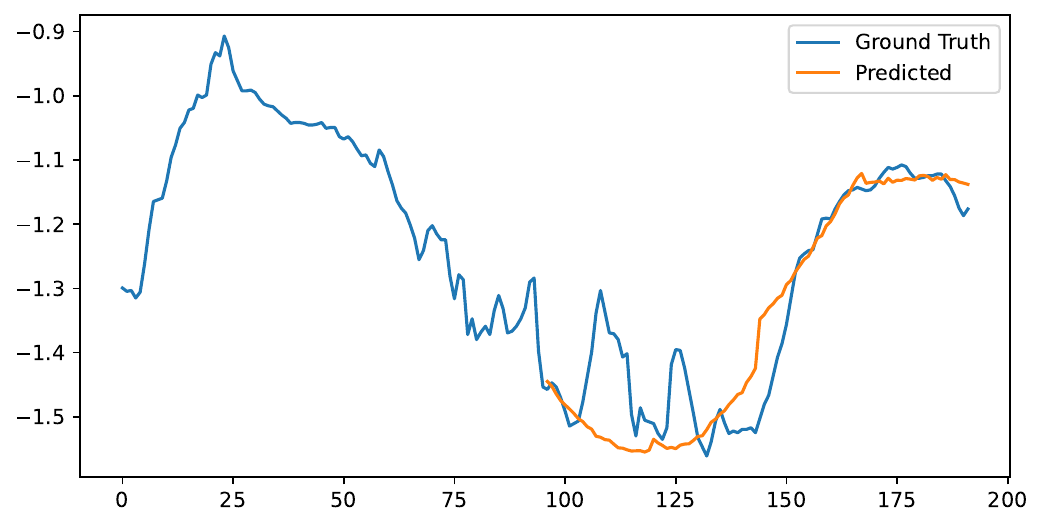}}
    % \quad
    \subfloat[{Averaged score}]{%
       \includegraphics[width=0.215\linewidth]{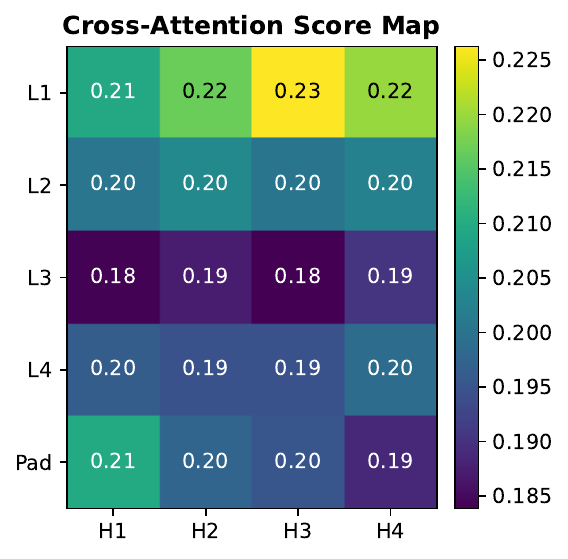}}
    \subfloat[{L1-H2}]{%
       \includegraphics[width=0.19\linewidth]{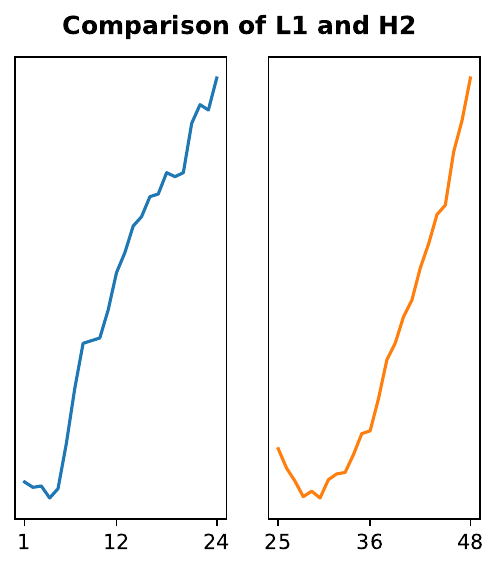}}
    \subfloat[{L1-H3}]{%
       \includegraphics[width=0.19\linewidth]{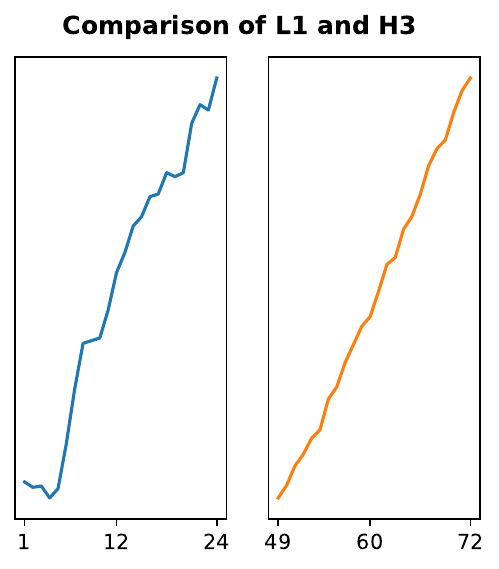}}
    % \quad
    \caption{Illustration of (a) forecasting result, (b) averaged cross-attention score, and (c,d) patches with the highest score on Weather. The score map is averaged from all the heads across layers.}
    \label{fig:real_score_map_weather}
\end{figure}

\begin{figure}[ht!] 
    \centering
    \subfloat[{Forecasting results}]{%
       \includegraphics[width=0.4\linewidth]{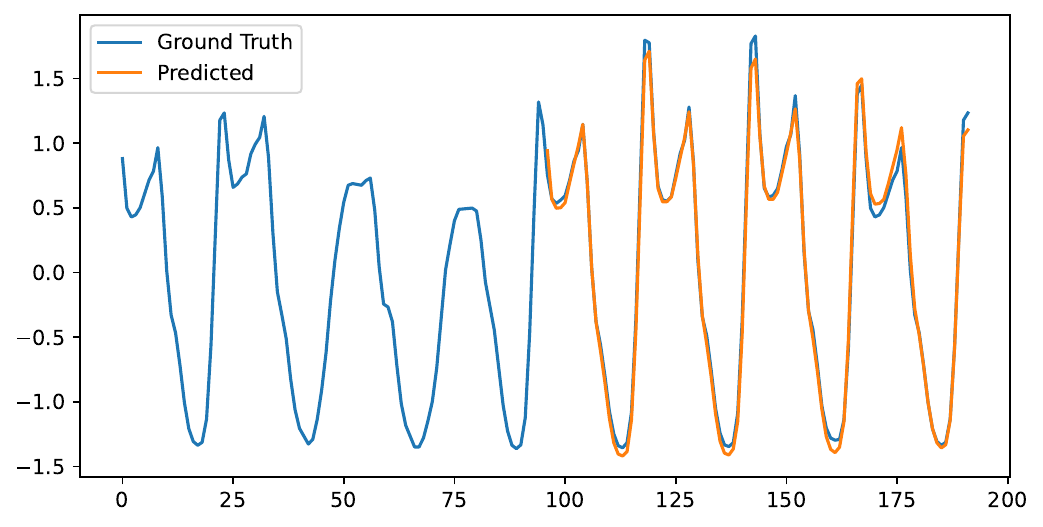}}
    % \quad
    \subfloat[{Averaged score}]{%
       \includegraphics[width=0.215\linewidth]{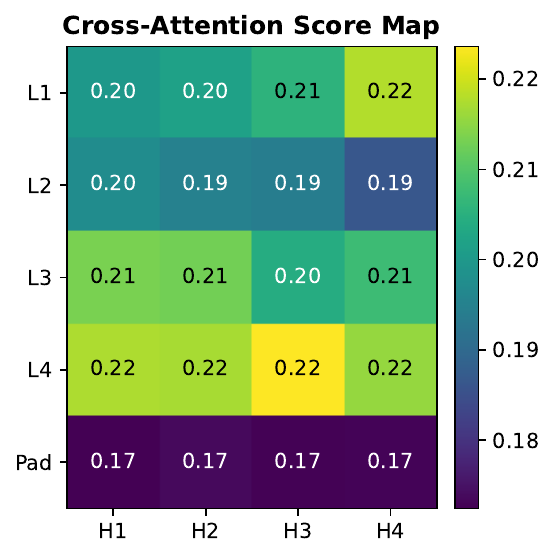}}
    \subfloat[{L4-H3}]{%
       \includegraphics[width=0.19\linewidth]{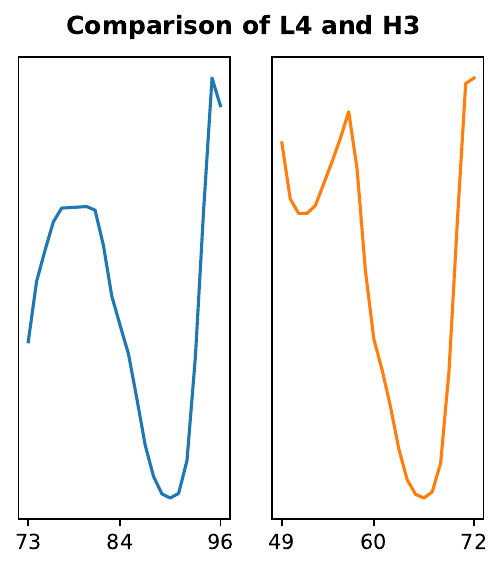}}
    \subfloat[{L1-H4}]{%
       \includegraphics[width=0.19\linewidth]{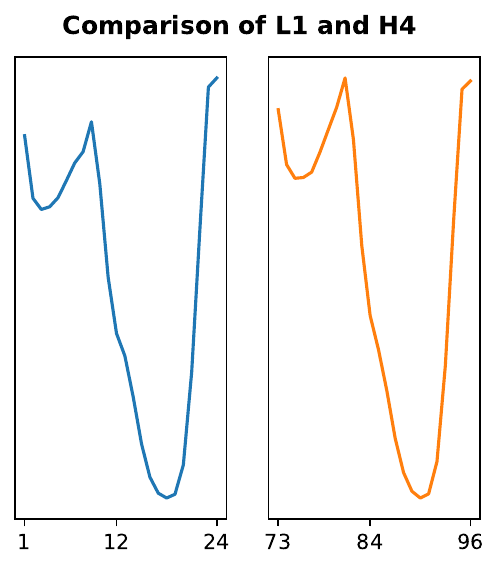}}
    % \quad
    \caption{Illustration of (a) forecasting result, (b) averaged cross-attention score, and (c,d) patches with the highest score on Traffic. The score map is averaged from all the heads across layers.}
    \label{fig:real_score_map_traffic}
\end{figure}

\begin{figure}[ht!] 
    \centering
    \subfloat[{Forecasting results}]{%
       \includegraphics[width=0.4\linewidth]{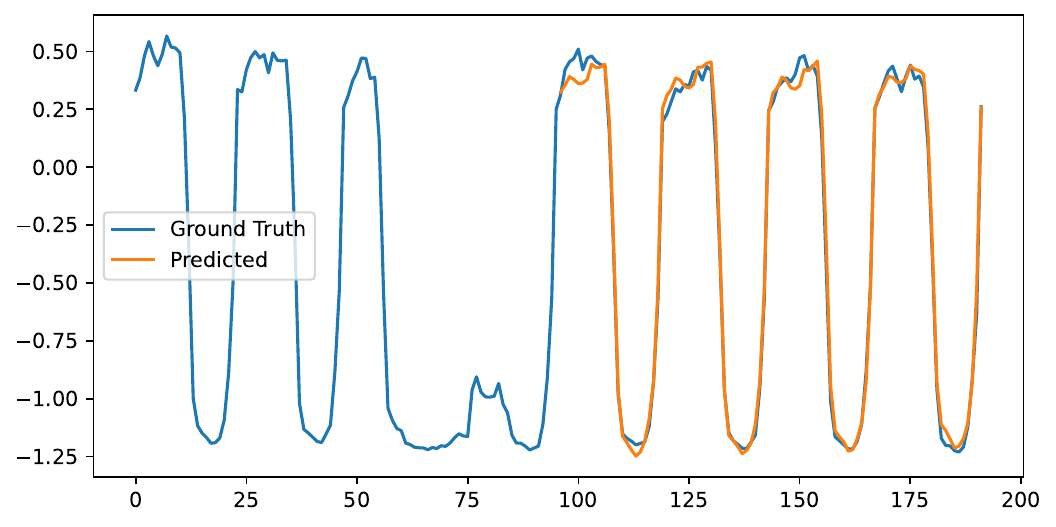}}
    % \quad
    \subfloat[{Averaged score}]{%
       \includegraphics[width=0.215\linewidth]{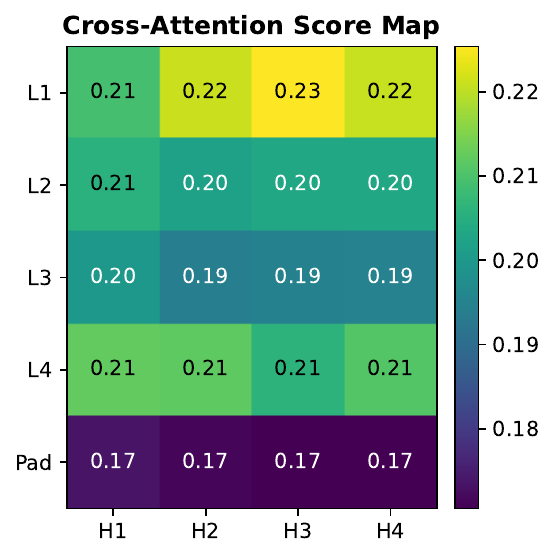}}
    \subfloat[{L1-H2}]{%
       \includegraphics[width=0.19\linewidth]{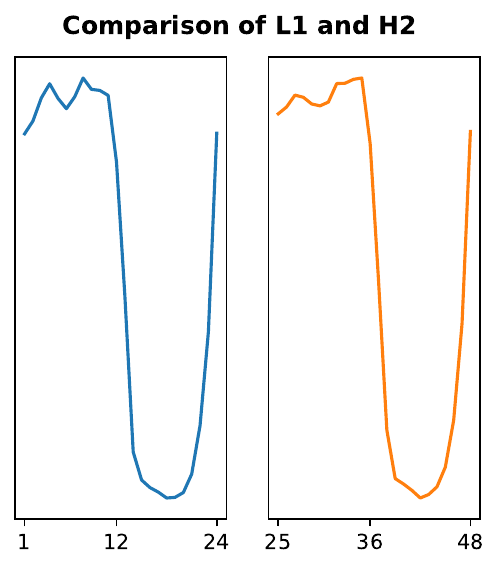}}
    \subfloat[{L1-H3}]{%
       \includegraphics[width=0.19\linewidth]{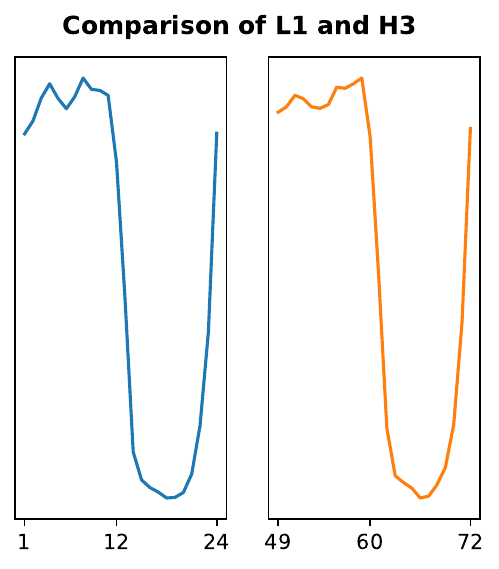}}
    % \quad
    \caption{Illustration of (a) forecasting result, (b) averaged cross-attention score, and (c,d) patches with the highest score on Electricity. The score map is averaged from all the heads across layers.}
    \label{fig:real_score_map_electricity}
\end{figure}

\begin{figure}[ht!] 
    \centering
    \subfloat[{Forecasting results}]{%
       \includegraphics[width=0.4\linewidth]{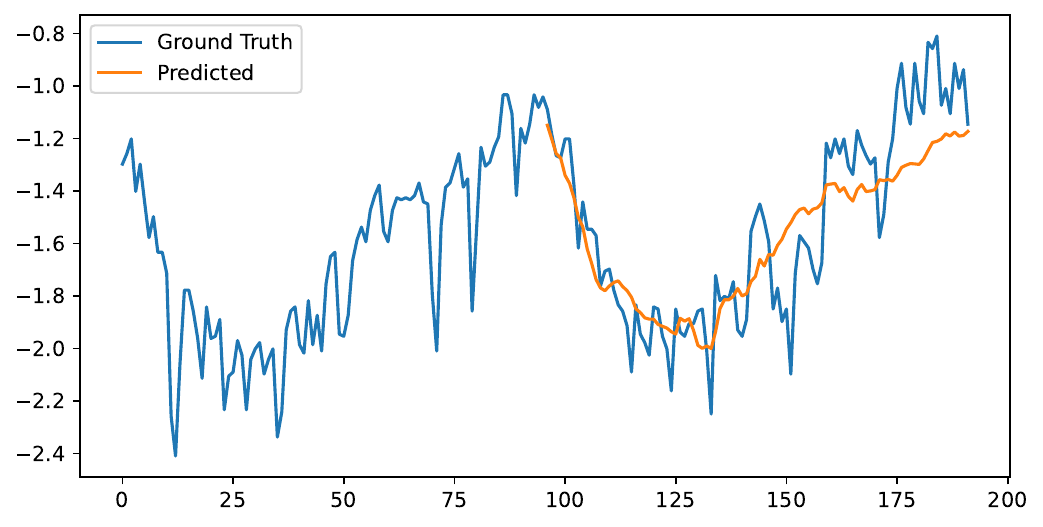}}
    % \quad
    \subfloat[{Averaged score}]{%
       \includegraphics[width=0.215\linewidth]{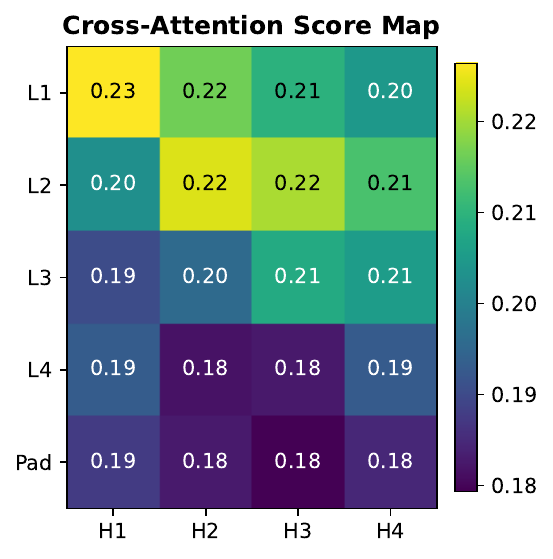}}
    \subfloat[{L1-H1}]{%
       \includegraphics[width=0.19\linewidth]{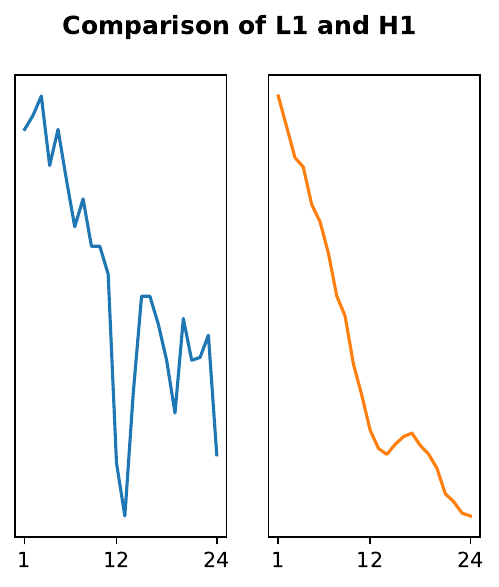}}
    \subfloat[{L2-H2}]{%
       \includegraphics[width=0.19\linewidth]{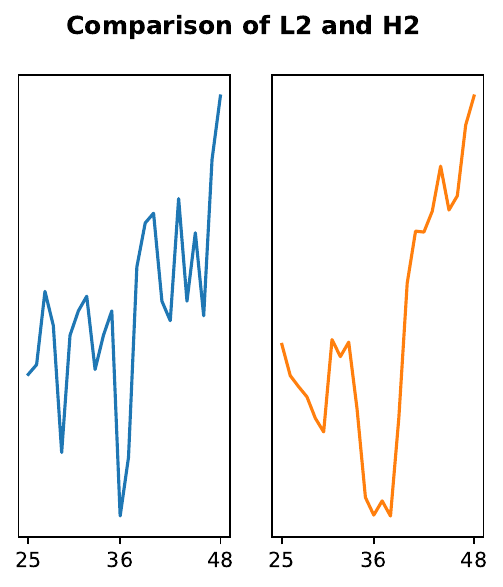}}
    % \quad
    \caption{Illustration of (a) forecasting result, (b) averaged cross-attention score, and (c,d) patches with the highest score on ETTm2. The score map is averaged from all the heads across layers.}
    \label{fig:real_score_map_ettm2}
\end{figure}

\begin{figure}[ht!] 
    \centering
    \subfloat[{Forecasting results}]{%
       \includegraphics[width=0.4\linewidth]{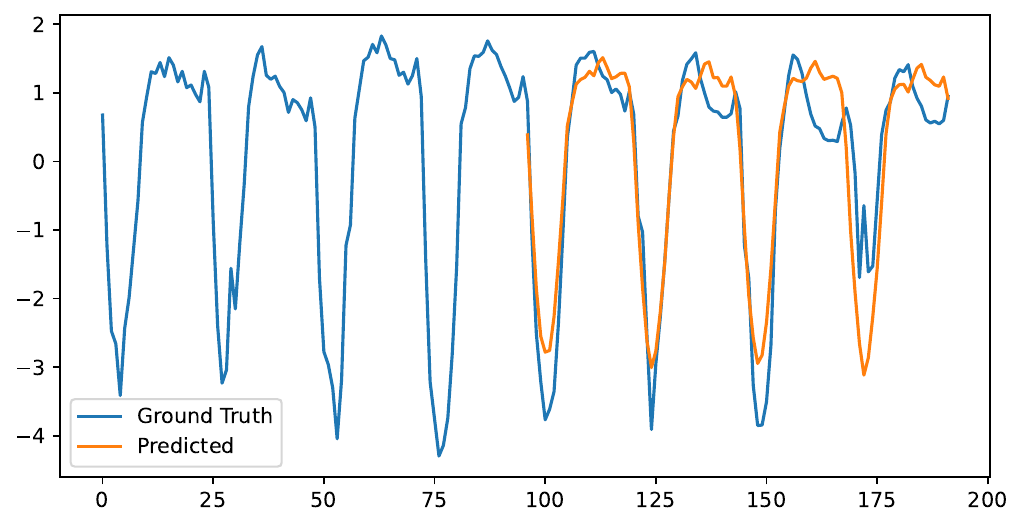}}
    % \quad
    \subfloat[{Averaged score}]{%
       \includegraphics[width=0.20\linewidth]{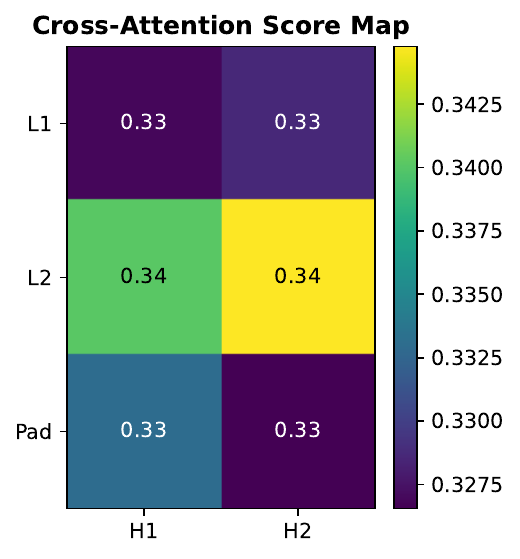}}
    \subfloat[{L2-H1}]{%
       \includegraphics[width=0.19\linewidth]{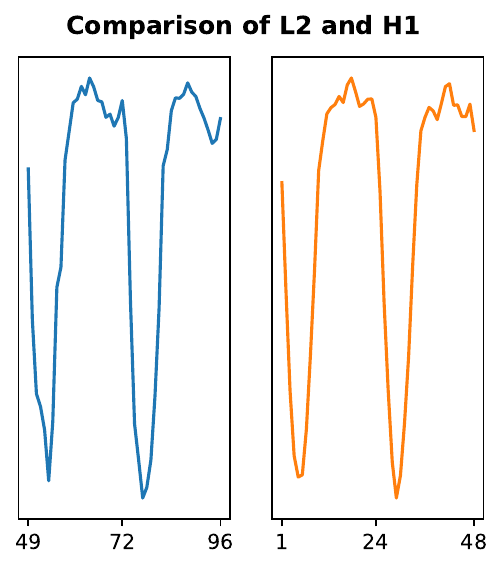}}
    \subfloat[{L2-H2}]{%
       \includegraphics[width=0.19\linewidth]{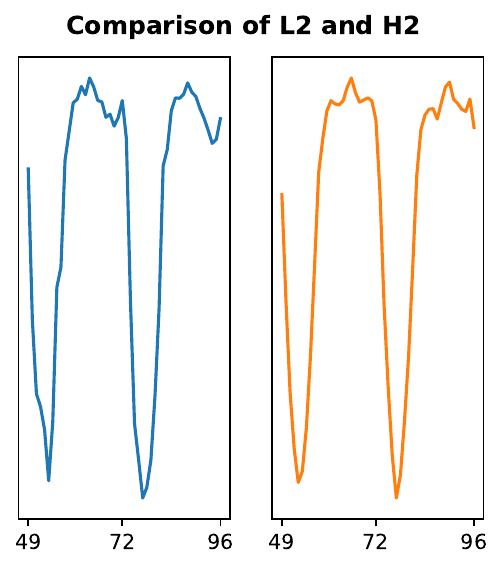}}
    % \quad
    \caption{Illustration of (a) forecasting result, (b) averaged cross-attention score, and (c,d) patches with the highest score on ETTh1. The score map is averaged from all the heads across layers.}
    \label{fig:real_score_map_etth1}
\end{figure}

\begin{figure}[ht!] 
    \centering
    \subfloat[{Forecasting results}]{%
       \includegraphics[width=0.4\linewidth]{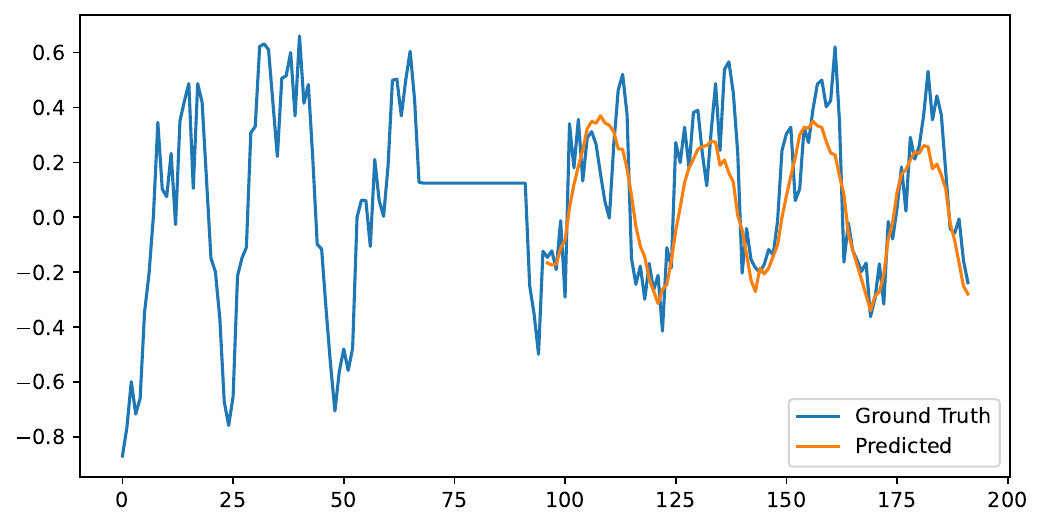}}
    % \quad
    \subfloat[{Averaged score} ]{%
       \includegraphics[width=0.19\linewidth]{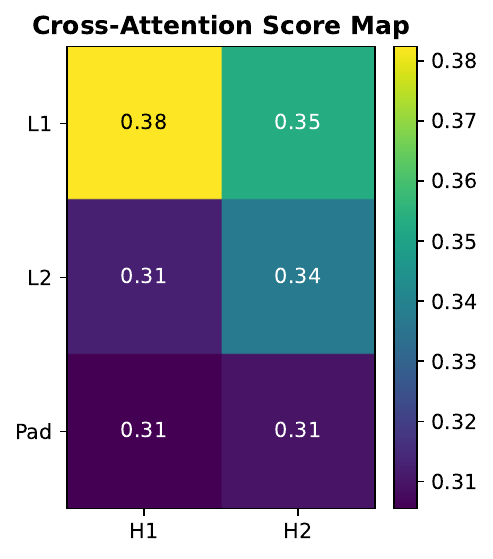}}
    \subfloat[{L1-H1}]{%
       \includegraphics[width=0.19\linewidth]{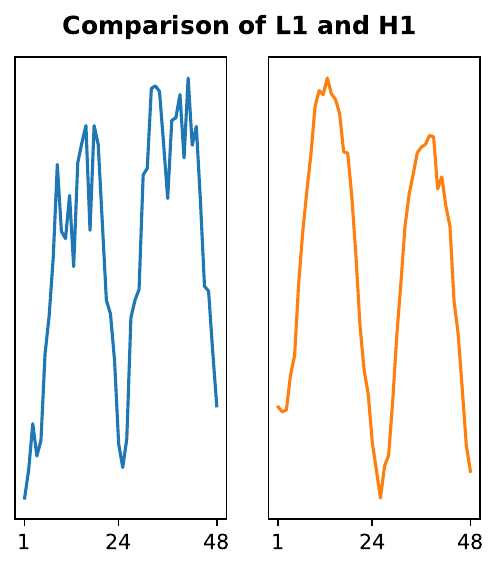}}
    \subfloat[{L1-H2} ]{%
       \includegraphics[width=0.19\linewidth]{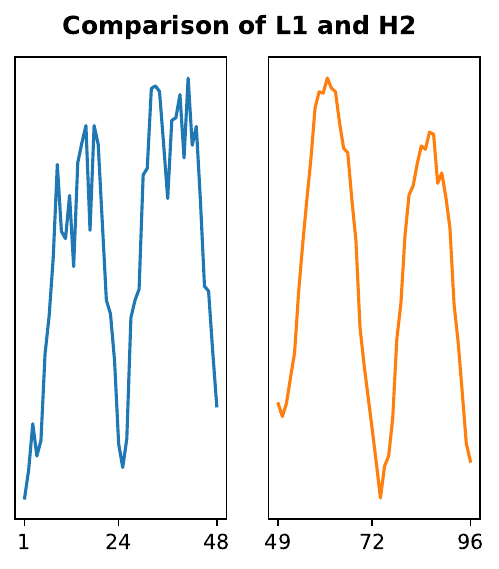}}
    % \quad
    \caption{Illustration of (a) forecasting result, (b) averaged cross-attention score, and (c,d) patches with the highest score on ETTh2. The score map is averaged from all the heads across layers.}
    \label{fig:real_score_map_etth2}
\end{figure}

To push further, we reproduce the experiment of \Figref{fig:real_score_map} with other datasets used in forecasting tasks. 
We illustrate the results of the Weather, Traffic, Electricity, ETTm2, ETTh1, and ETTh2 in Figures \ref{fig:real_score_map_weather}, \ref{fig:real_score_map_traffic}, \ref{fig:real_score_map_electricity}, \ref{fig:real_score_map_ettm2}, \ref{fig:real_score_map_etth1}, and \ref{fig:real_score_map_etth2}, respectively. 
For each figure, (a) represents the forecasting results, (b) shows the cross-attention score map, and (c) and (d) illustrate the two pairs with the highest attention scores. For all figures, our attention-based explanation successfully discovers similar periodic patterns. Therefore, we believe that our model has the potential to provide a clearer understanding of the mechanisms underlying forecasting predictions. We hope that future research will continue to explore and expand upon this foundation.

\end{document}